\documentclass[11pt]{article}

\usepackage{deauthor,times,graphicx}
\usepackage[square,comma,numbers,sort]{natbib}
\usepackage{url}
\usepackage{hyperref}
\usepackage{multirow,multicol,xspace}
\usepackage{amsmath,amssymb}
\usepackage{float}
\usepackage{booktabs}
\usepackage{graphics}
\usepackage{enumitem}
\usepackage{amsmath}
\usepackage{lipsum}
\usepackage{makecell}

\usepackage{pifont}
\usepackage[edges]{forest}
\usepackage{xcolor}
\usepackage[most]{tcolorbox}
\usepackage{color}
\definecolor{lighterseafoam}{RGB}{242,245,224}

\newtcolorbox{mybox}{
    colback=lighterseafoam, 
    boxrule=1pt,
    width=\textwidth,
    breakable
}

\newtcolorbox{promptbox}{
    colback=lighterseafoam, 
    boxrule=1pt,
    width=1\textwidth,
    breakable
}
\definecolor{codeblue}{rgb}{0.25,0.5,0.5}
\hypersetup{
    colorlinks=true,
    citecolor=codeblue,
    bookmarksnumbered=true,
    pdfpagemode={UseOutlines},
    plainpages=false,
    pdftitle={User Modeling in the Era of Large Language Models},
    pdfauthor={Z. Tan, et al.}
}

\renewcommand\paragraph[1]{\vspace{0.05in} \noindent \textbf{#1.}}

\bulletinyear{2023}

\begin{document}
\title{User Modeling in the Era of Large Language Models: \\Current Research and Future Directions}
\author
{Zhaoxuan Tan, Meng Jiang \\
\small{Department of Computer Science and Engineering, University of Notre Dame}\\ 
\small\texttt{\{ztan3, mjiang2\}@nd.edu}
}
\maketitle

\begin{abstract}
User modeling (UM) aims to discover patterns or learn representations from user data about the characteristics of a specific user, such as profile, preference, and personality. The user models enable personalization and suspiciousness detection in many online applications such as recommendation, education, and healthcare. Two common types of user data are text and graph, as the data usually contain a large amount of user-generated content (UGC) and online interactions. The research of text and graph mining is developing rapidly, contributing many notable solutions in the past two decades. Recently, large language models (LLMs) have shown superior performance on generating, understanding, and even reasoning over text data. The approaches of user modeling have been equipped with LLMs and soon become outstanding. This article summarizes existing research about how and why LLMs are great tools of modeling and understanding UGC. Then it reviews a few categories of large language models for user modeling (LLM-UM) approaches that integrate the LLMs with text and graph-based methods in different ways. Then it introduces specific LLM-UM techniques for a variety of UM applications. Finally, it presents remaining challenges and future directions in the LLM-UM research. We maintain the reading list at: \url{https://github.com/TamSiuhin/LLM-UM-Reading}.


\end{abstract}

\section{Introduction}
\label{sec:intro}

\begin{figure}[h]
  \centering
  \includegraphics[width=0.75\textwidth]{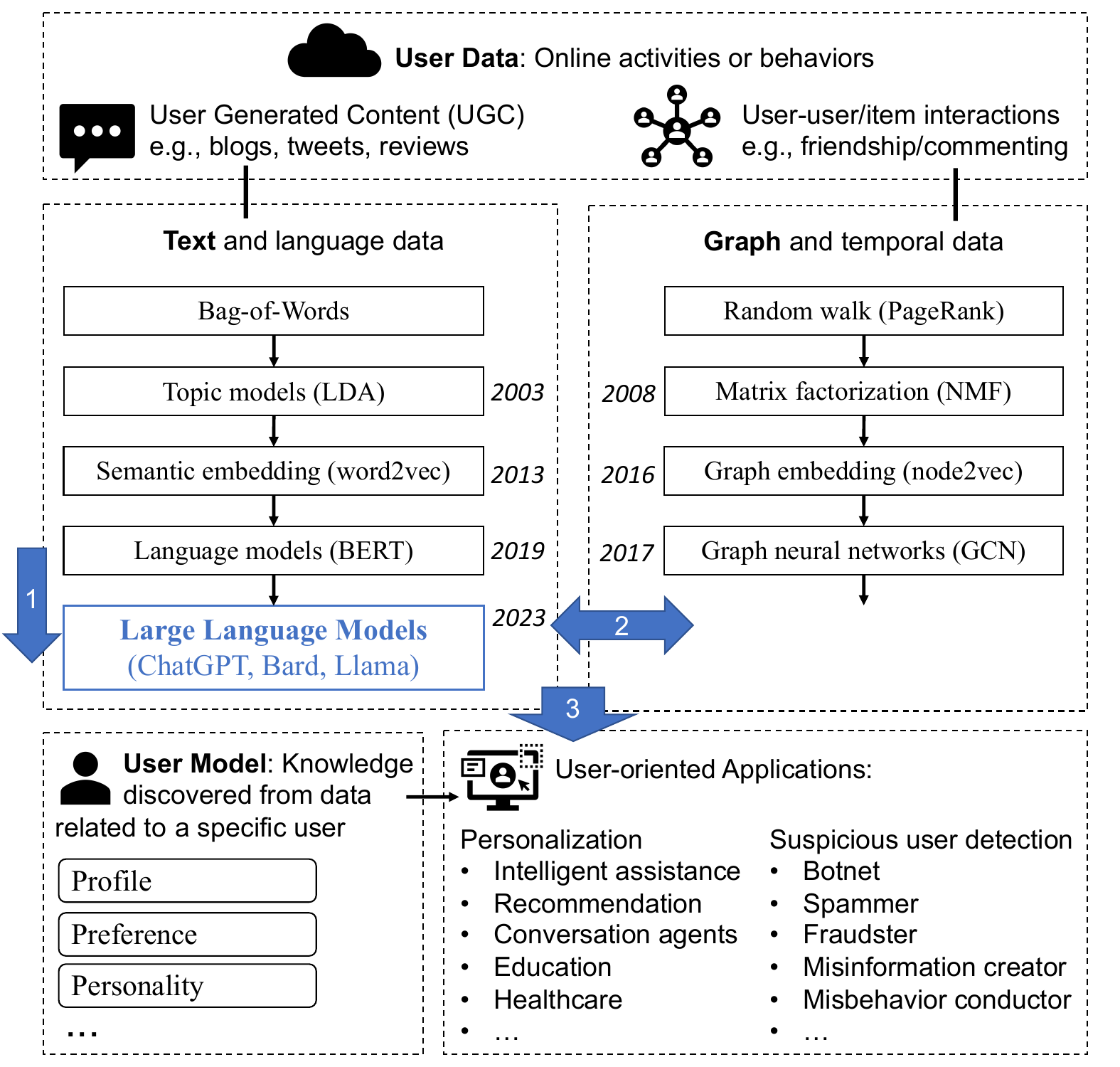}
  \caption{User modeling aims to discover knowledge and patterns from user data to identify profile, preference, and personality. The three blue arrows in the figure correspond to our three major contributions: (1) summarize how and why LLMs are great tools for modeling and understanding UGC, (2) review approaches that integrate LLMs with text- and graph-based UM methods, and (3) introduce LLM-UM techniques for various applications.}
  \label{fig:intro}
\end{figure}


User Modeling (UM) aims to extract valuable insights and patterns from user behaviors, enabling customization and adaptation of systems to meet specific users' needs \cite{li2021survey}. UM techniques have facilitated a better understanding of user behaviors, customized intelligent assistance, and greatly improved user experience. For example, when people are looking for dinner options and searching online, the UM techniques infer their characteristics based on interaction history, predict current food interests, and give personalized recommendations. UM has a substantial impact on user data analysis and many applications such as e-commerce \cite{schafer2001commerce,zhao2015commerce, sarwar2000analysis}, entertainment \cite{berkovsky2005entertainment, christensen2011entertainment, natkin2006user}, and social networks \cite{abel2011analyzing, tao2012tums, abel2013twitter}. UM is a highly active and influential research field.


User modeling is mainly about mining and learning user data, including user-generated content (UGC) and user's interactions with other users and items.
User-generated content encompasses a wide range of text data, such as tweets, reviews, blogs, and academic papers. The rich texts can be analyzed by natural language processing (NLP) techniques. User interactions, on the other hand, involve various actions such as following, sharing, rating, comments, and retweets. These interactions may form a heterogeneous temporal text-attributed graph \cite{shi2016survey} for having temporal and textual information and having different types of nodes and relations. That can be analyzed using graph mining and learning techniques. As a result, user modeling has branched out into text-based and graph-based approaches, focusing on extracting insights from text and graph data, respectively.

\vspace{0.05in}
\noindent \textbf{How has the research of text-based UM developed?} Researchers have used multiple types of representations of texts, such as words, topics, and embeddings. 
Bag-of-Words (BoW) models create distributional text representations with word frequencies using a discrete vocabulary \cite{harris1954distributional}.
To address the sparsity of BoW representations, topic modeling techniques statistically discover latent topics in a collection of documents, e.g., latent Dirichlet allocation (LDA) \cite{blei2003latent}.
But they are not able to capture semantic meanings, i.e., word semantic similarity.
Word2Vec employs nonlinear neural layers to develop Continuous Bag-of-Words (CBOW) and continuous skip-gram models \cite{mikolov2013efficient}. It extracts semantic embeddings from many kinds of UGC text data, such as blogs, reviews, and tweets.
However, the neural layers are too shallow to capture deep sequential patterns among numerous word tokens.
With the breakthrough in Transformer architectures \cite{vaswani2017attention}, pretrained language models (PLMs) significantly changed the landscape of UGC understanding with the pretrain-fine-tune paradigm. The new paradigm trains the models on a large unlabeled corpus using self-supervised learning and uses hundreds or thousands of examples to fine-tune the models for downstream task adaptation \cite{kenton2019bert}. Recently, large language models (LLMs) have revolutionized this area with emergent abilities, including unprecedented reasoning \cite{wei2022chain, yao2023tree}, generalization \cite{wei2021finetuned, radford2019language}, and knowledge comprehension \cite{sun2023head,pan2023unifying}.
LLMs are operated under the pretraining paradigm on extremely large corpora to update billions of parameters. A lot of research has shown that LLMs understand UGC in a zero-shot manner, i.e., without a collection of examples for fine-tuning. LLMs have surpassed human performance in summarization \cite{pu2023summarization}, outperformed most human in several exams \cite{openai2023gpt4}, and shown strong reasoning abilities with prompt engineering, including Chain-of-Thought \cite{wei2022chain}, Least-to-Most \cite{zhou2022least}, and Tree-of-Thought \cite{yao2023tree}. LLMs open a new era for UM research to re-think about UGC mining.

\vspace{0.05in}
\noindent \textbf{How has the research of graph-based UM developed?} User interactions with online content and users are naturally defined as edges that connect nodes of users or things. The user data can be defined as a graph. Heterogeneous graphs contain multiple types of nodes (e.g., users, items, places) and relations. Temporal/weighted graphs have timestamps/weights labeled on the interactions. Attributed graphs allow nodes to have a set of attribute-value pairs (e.g., age of a user, color of a product). In text-rich graphs, the nodes have long-form text attributes. Random walk with restarts provides a closeness score between two nodes in a weighted graph, and it has been successfully used in numerous settings (e.g., personalized PageRank \cite{page1998pagerank}).
Matrix factorization (MF) decomposes the user-item interaction matrix into the product of two matrices or known as latent features of users and items \cite{koren2009matrix,jiang2012social,jiang2014scalable}. Regarding collaborative filtering, MF performs better with explicit feedback ratings, while RWR exploits the global popularity of items. It is actually a basic embedding model \cite{zhang2018network}.
With deep learning, Node2Vec extracts sequences from the graph with random walks and uses Word2Vec to learn node embeddings \cite{grover2016node2vec}. However, encoding a graph into sequences would cause information loss. Graph neural networks (GNNs) employ the message-passing mechanism for deep representation learning on graphs. Specifically, the family of Graph Convolutional Network (GCN) \cite{kipf2016semi} has greatly improved the performance in recommendation \cite{fan2019graph, he2020lightgcn}, user profiling \cite{chen2019semi, yan2021relation}, user behavior prediction \cite{yu2020identifying, wang2020calendar}, and suspicious user detection \cite{feng2022twibot, feng2022heterogeneity}. 

\vspace{0.05in}
\noindent \textbf{Why are LLMs revolutionizing text- and graph-based UM research?} 
User modeling involves a series of machine learning tasks on text and graph data, such as text classification, node classification, link prediction, and time series modeling. Putting into context, the tasks can be sentiment analysis, natural language inference (NLI), user and product categorization, social relationship prediction, and temporal behavior prediction. Traditionally, the solution must be a specific model for a specific type of data and be trained on a specific set of annotations. For example, due to schema differences, two text classifiers had to be trained for the sentiment analysis and NLI tasks separately. Also, two networks or at least two modules in a graph neural network (GNN) were trained to predict if a user makes a new friend and purchases an item, respectively. Moreover, the textual information of user and/or product profiles is quite limited for learning and prediction due to the long-tail distribution.

LLMs have changed the paradigm of solution development. First, if designed properly, the prompts are able to treat most of the text-to-label tasks in LLMs as a unified text generation task; annotation data become not desperately needed; and the performance can even be comparable to or better than the traditional models. This is because LLMs were pre-trained on extremely large corpora and fine-tuned to follow the instructions in the prompts. Second, the prompts can be designed for learning tasks on graph data. For example, one can ask LLMs ``if a user bought an Apple watch yesterday, will the user consider buying a pair of running shoes?'' The ``analysis'' by LLMs can provide additional information to existing user-item link predictors. Third, all text information can be automatically expanded by LLMs. The relevant parametric knowledge augments the input of machine learning models and reduces the task difficulty.

LLMs have showcased robust capabilities in characterizing users' personalities \cite{rao2023can}, discerning users' stances \cite{zhang2022would}, pinpointing user preferences \cite{fang2023chatgpt}, and beyond. Also, they have demonstrated marked proficiency in node classification \cite{ye2023natural}, node attribute prediction \cite{he2023explanations}, and graph reasoning \cite{wang2023can}. Preliminary research focuses on leveraging LLMs for user modeling (LLM-UM) to integrate text-based and graph-based methods. For user profiling, GENRE \cite{liu2023genre} leverages ChatGPT as a user profiler by feeding users' behavior history and prompting the model to infer the users' preferred topics and regions. These LLM-generated profiles serve as important features for click-through rate recommendation models and resolve the anonymity problem in collecting user profiles. For recommendation, \citet{kang2023llms} use LLMs to predict user ratings based on their behavior history and find that LLMs typically demand less data while maintaining world knowledge about humans and items. For personalization, LaMP \cite{salemi2023lamp} proposes a benchmark incorporating personalized text generation and classification tasks as well as a retrieval-augmented approach. LLMs can be personalization tools for their understanding of user data. On suspiciousness detection, \citet{chiu2021detecting} employ GPT-3 to detect hate speech, discovering that LLMs are able to identify abusive language with limited labels.

\paragraph{Difference from existing surveys}
Given the growing interest and expanding body of work in user modeling (UM) with LLMs, this is an ideal opportunity to provide a comprehensive review that accomplishes several goals. In this survey, we analyze the advantages of LLMs in boosting existing user modeling techniques, introduce a taxonomy that categorizes LLM-UM techniques from the perspective of methodology, provide an in-depth review of specific techniques for a wide range of real-world applications, and finally outline the challenges and potential future directions in the field. We aim to provide a handbook for researchers and practitioners in the relevant fields, so they are able to confidently use LLMs to design and develop effective UM approaches.

It is worthwhile to discuss the uniqueness of our effort.  \citet{farid2018user} conducted a survey on user profiling approaches as a part of user modeling before LLMs came out. \citet{li2021survey} reviewed representation learning methods for user modeling, which was a prevailing paradigm before the advent of LLMs. \citet{he2023survey} focused on user behavior modeling within recommender systems, narrowing down the scope specifically to item recommendations in the pre-LLM era.
In the context of the LLM era, \citet{fan2023recommender, wu2023survey} investigated the applications of LLMs in recommender systems. Additionally, \citet{lin2023can} examined approaches that incorporated LLMs to enhance recommender systems, while \citet{chen2023large} summarized recent work, challenges, and future directions in LLM-based personalization. However, these surveys discussed specific application goals of user modeling, either recommendation or personalization, which represents only a small portion of user modeling. To the best of our knowledge, there is no survey that summarizes the LLM-UM research work. Hence, our survey aims to fill this gap by providing a comprehensive summary and inspiring future research directions.

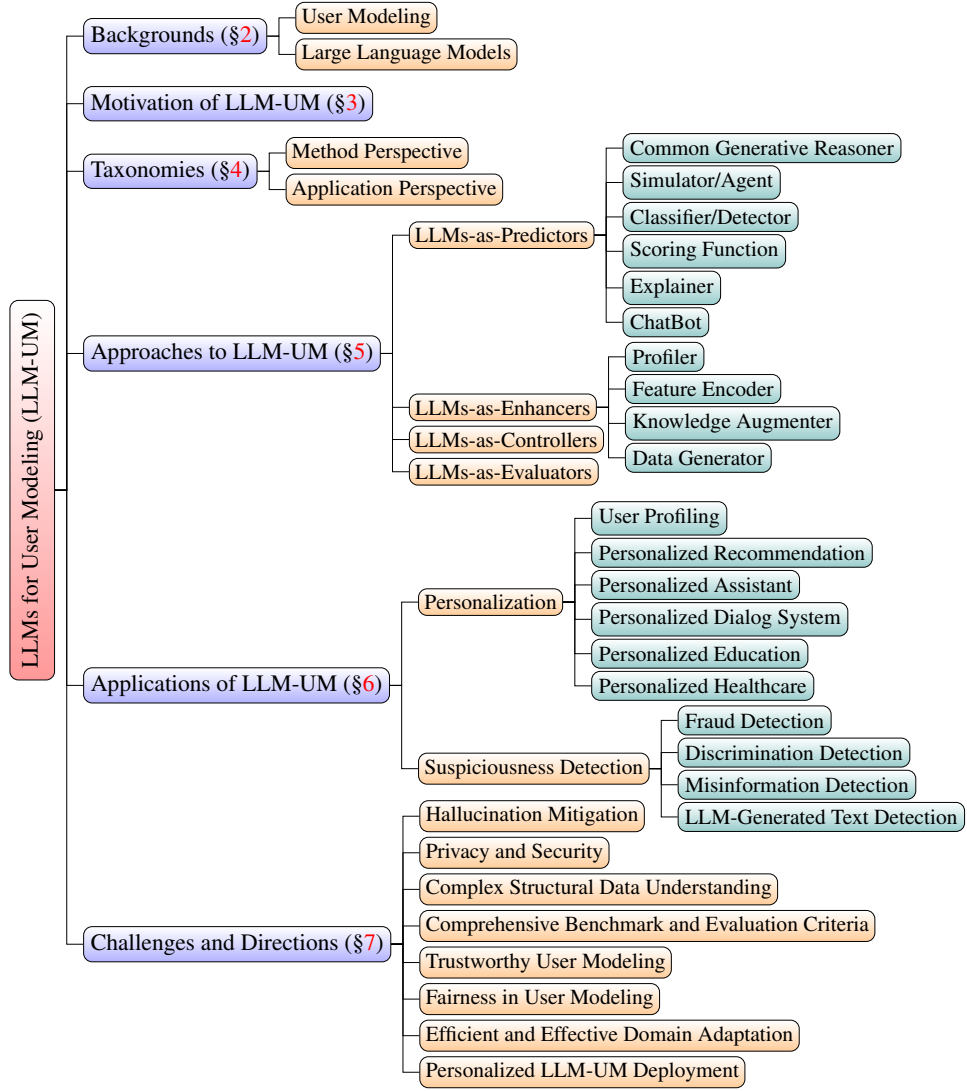
\begin{figure}[!t]
\centering
\resizebox{0.75\linewidth}{!}{
\begin{forest}
	for tree={
		draw,
		shape=rectangle,
		rounded corners,
		top color=white,
		grow'=0,
		l sep'=1.2em,
		reversed=true,
		anchor=west,
		child anchor=west,
	},
	forked edges,
	root/.style={
		rotate=90, shading angle=90, bottom color=red!40,
		anchor=north, font=\normalsize, inner sep=0.5em},
	level1/.style={
		bottom color=blue!30, font=\normalsize, inner sep=0.3em,
		s sep=0.2em},
	level2/.style={
		bottom color=orange!40, font=\small, inner sep=0.25em,
		s sep=0.1em},
	level3/.style={
		bottom color=teal!40, font=\small, inner sep=0.3em,
		l sep'=0.5em},
	where n=0{root}{},
	where level=1{level1}{},
	where level=2{level2}{},
	where level=3{level3}{},
	[LLMs for User Modeling (LLM-UM)
            [Backgrounds (\S \ref{sec:background})
                [User Modeling]
                [Large Language Models]
            ]
            [Motivation of LLM-UM (\S \ref{sec:motivation})]
		[Taxonomies (\S \ref{sec:taxonomy})
                [Method Perspective]
                [Application Perspective]
		]
            [Approaches to LLM-UM (\S \ref{sec:approach})
                [LLMs-as-Predictors
                    [Common Generative Reasoner]
                    [Simulator/Agent]
                    [Classifier/Detector]
                    [Scoring Function]
                    [Explainer]
                    [ChatBot]
                ]
                [LLMs-as-Enhancers
                    [Profiler]
                    [Feature Encoder]
                    [Knowledge Augmenter]
                    [Data Generator]
                ]
                [LLMs-as-Controllers]
                [LLMs-as-Evaluators]
		]
		[Applications of LLM-UM (\S \ref{sec:application})
                [Personalization
                    [User Profiling]
                    [Personalized Recommendation]
                    [Personalized Assistant]
                    [Personalized Dialog System]
                    [Personalized Education]
                    [Personalized Healthcare]
                ]
                [Suspiciousness Detection
                    [Fraud Detection]
                    [Discrimination Detection]
                    [Misinformation Detection]
                    [LLM-Generated Text Detection]
                ]
		]
            [Challenges and Directions (\S \ref{sec:future})
                [Hallucination Mitigation]
                [Privacy and Security]
                [Complex Structural Data Understanding]
                [Comprehensive Benchmark and Evaluation Criteria]
                [Trustworthy User Modeling]
                [Fairness in User Modeling]
                [Efficient and Effective Domain Adaptation]
                [Personalized LLM-UM Deployment]
            ]
	]
\end{forest}
}
\caption{Structure of this survey.}
\label{fig:overview}
\end{figure}

The remainder of this survey is organized as follows (Figure \ref{fig:overview}). Section \ref{sec:background} gives the background of user modeling techniques and large language models and gives the motivation of why LLMs are good tools for next-generation user modeling. Section \ref{sec:taxonomy} introduces two taxonomies of LLM-UM based on their approaches and applications. Section \ref{sec:approach} summarizes the approaches to LLM-UM and how LLMs can integrate text and graph-based methods in existing works, including leveraging LLMs as enhancers, predictors, and controllers. Section \ref{sec:application} elaborates on the LLM-UM applications, including personalization and suspiciousness detection. Finally, Section \ref{sec:future} delves into current challenges and future directions pertaining to the LLM-UM topic.



\section{Background}
\label{sec:background}

\subsection{User Modeling}



User modeling (UM) involves insight extraction or prediction from user data, such as profiles, personality traits, behavior patterns, and preferences. The insights can be utilized to customize and optimize user-oriented systems or services, enabling them to adapt effectively to the unique requirements of individual users \cite{li2021survey}.
From a user data standpoint, there are two primary types: user-generated content (UGC) and user-user/item interaction. These data modalities encompass textual content and graph-based interactions. User modeling techniques can be broadly classified into two categories: text-based methods and graph-based methods, each with a distinct emphasis on UGC and user interaction graphs, respectively.


\paragraph{Text-based UM Methods} The text-based UM methods focus on mining user-generated content, understanding user profiles, personalities, and subsequently inferring user preference and providing personalized recommendations and assistance. Text-based UM methods are highly relevant to natural language processing and have experienced several breakthroughs in the last two decades. 

Initially, people widely use Bag-of-Words (BoW) \cite{harris1954distributional} that maintains a discrete word vocabulary and represents text as an unordered collection of words. However, BoW disregards grammar, word order, and semantics. Later, topic models are proposed to identify groups of similar words in an unsupervised statistical machine learning manner \cite{alghamdi2015survey}. The most representative topic model is Latent Dirichlet Analysis (LDA) \cite{blei2003latent}, which considers documents as a mixture of topics and topics as a mixture of words. Latent semantic analysis (LSA) \cite{dumais2004latent} is another popular topic model that is based on the principle that words close in meaning tend to be used together in context.
Soon after, \citet{mikolov2013efficient} introduced neural network techniques for UGC mining and understanding, including the continuous Bag-of-Words (CBOW) that predicts current words based on the context and continuous skip-gram model that predicts surrounding words given the current words. The emergence of Word2Vec brings new ideas to user modeling; for example, \citet{hu2017user} use Word2Vec to encode user search history and predict users' age, gender, and education. However, the simplicity of Word2Vec leads to a limited semantic understanding of UGC. The emergence of Transformer architecture \cite{vaswani2017attention} and pretrained language models (PLMs) \cite{kenton2019bert,qiu2020pre} dramatically change the landscape of natural language processing, resulting in significant progress in text-based user modeling. For instance, BERT4Rec \cite{sun2019bert4rec} utilizes a bidirectional self-attention mechanism to learn user behavior sequence representations for recommendation. Recently, large language models (LLMs) and their unprecedented emergent ability lead to a revolution in text-based user modeling. Specifically, LLMs are trained on large amounts of textual data with billions of learnable parameters to understand the patterns, semantics, and structure of natural language. Some preliminary explorations have shown LLMs are instrumental in enhancing functionality and adapting to users' specific needs in user modeling systems. LLMs have the capability to serve as a user profiler to generate user characteristics and personality, as in representative work such as GENRE \cite{liu2023genre}, PALR \cite{chen2023palr}, and MBTI-based assessment \citet{rao2023can}. Furthermore, LLMs are proven to be powerful recommender systems, evidenced by LLMRec \cite{liu2023llmrec}, Chat-REC \cite{gao2023chat}, and LKPNR \cite{runfeng2023lkpnr}. LLM personalization has also been a prominent area of research, delivering products such as LaMP \cite{salemi2023lamp}, AuthorPred \cite{li2023teach}, and NetGPT \cite{chen2023netgpt}. Additionally, several studies have explored the utilization of LLMs for the detection of misinformation and misbehavior \cite{mozes2023use, labonne2023spam}. 






\paragraph{Graph-based UM Methods} The graph-based user modeling methods focus on learning from graph structures about user-user/item interactions. Moreover, the timestamp and intrinsic heterogeneity of interaction networks enrich the graph information, together forming a temporal heterogeneous graph. To mine the graph and temporal data for user modeling, \citet{page1998pagerank} first propose PageRank, which measures the importance of a node based on times visited by the random walk. Collaborative filtering \cite{schafer2007collaborative} is then proposed, assuming that users with similar behaviors would rate and act on items in a similar manner. Matrix factorization, as a collaborative filtering technique, has dominated the user modeling field for quite a few years since Netflix competition for its scalability and flexibility \cite{singh2008relational, koren2009matrix, lee2000algorithms}. In its basic form, matrix factorization features both items and users in latent space based on user-item interaction history. High correspondence between the item and the user leads to a recommendation. Later, deep learning models are introduced to learn high-quality user and item embeddings in latent spaces \cite{perozzi2014deepwalk,grover2016node2vec,tang2015line,dong2017metapath2vec}. Node2Vec \cite{grover2016node2vec} is the pioneering work that uses random walks on social networks to generate sentences from graph structures and then feeds them into a Word2Vec model \cite{mikolov2013efficient} for graph embedding learning. Metapath2Vec \cite{dong2017metapath2vec} extends Node2Vec to heterogeneous graphs. To mitigate Node2Vec's information loss by extracting graphs into sequences, Graph Neural Network (GNN) is applied to advance user modeling with their robust structural encoding capabilities \cite{kipf2016semi, velivckovic2018graph, hu2020heterogeneous}. The core of graph neural networks lies in the message-passing mechanism that propagates node representations and aggregates neighborhood representations. The most representative GNN is graph convolutional network (GCN), which aggregates the information of different neighbors equally. Then the graph attention network (GAT) uses an attention mechanism to learn the attention weights of neighborhoods for higher-quality node representations. 
Specifically in user modeling applications, GNN-based methods are widely adopted in advanced user modeling systems and have achieved state-of-the-art performance. For instance, \citet{ying2018pinsage} design an efficient random walk algorithm, merge it with a type of GNN named GraphSAGE, and deploy the system on Pinterest.

\subsection{Large Language Model}
Language models are probabilistic models of natural language that can generate the likelihood of word sequences so as to predict the probabilities of future tokens \cite{lipretrained, zhao2023survey}. 
Large language models (LLMs) refer to the deep neural language models with billions of learnable parameters that are pretrained on an extremely large textual corpus to understand the distribution and structure of natural language \cite{zhao2023survey}. Thanks to the efficiency of the Transformer architecture \cite{vaswani2017attention}, almost all large language models employ it as the backbone. There are three types of language model design: encoder-only (\emph{e.g.}, BERT \cite{kenton2019bert}), decoder-only (\emph{e.g.}, GPT \cite{radford2018improving}), and encoder-decoder (\emph{e.g.}, T5 \cite{raffel2020t5}). Encoder-only models, specifically for BERT, use bidirectional attention to process token sequences and are pre-trained on masked token prediction and next-sentence classification tasks. This process can extract semantic embeddings for general purposes and enable the models to quickly adapt to diverse downstream tasks after fine-tuning. Decoder-only models, such as GPT, conduct text-to-text tasks based on the transformer decoder architecture. They are trained on the next token prediction tasks from left to right generation. Encoder-decoder models, such as T5, are trained on text-to-text tasks. Their encoders extract contextual representations from the input sequence, and their decoders use cross attention to map latent representations back to the text output space. In the context of LLMs, most models follow the decoder-only architecture as it simplifies the model and makes efficient inferences \cite{wang2022language}.




Recently, researchers have found that scaling pretrained language models' training data and parameter size often leads to a significant performance gain, \emph{a.k.a} scaling law \cite{kaplan2020scaling}. The large language models present emergent abilities \cite{wei2022emergent}, referring to the abilities that are not present in small models. Typically, there are three types of well-studied emergent abilities: in-context learning (ICL), instruction following, and step-by-step reasoning. In-context learning assumes that the language model has been provided with natural language instructions and/or several task demonstrations. LLMs can generate the expected output for test instances by completing the word sequence of input text without requiring additional training or gradient update, which is first introduced by GPT-3 \cite{brown2020language}. Recent ICL research focuses on reducing inductive bias \cite{levine2021inductive,si2023measuring}. The instruction following ability means that LLMs are shown to perform well on unseen tasks that are also described in the form of instructions after fine-tuning with a mixture of multi-task datasets formatted via natural language descriptions, known as instruction tuning. Instruction tuning improves the generalization ability of LLMs. The LLMs are better aligned with human intentions \cite{wang2023aligning}. Recent instruction tuning studies focus on how to align LLMs with tasks and user preferences effectively \cite{ouyang2022training} and efficiently \cite{zhou2023lima}. Step-by-step reasoning means that LLMs can solve complex tasks that require multi-step reasoning. Chain-of-Thought (CoT) \cite{wei2022chain} introduces intermediate steps of reasoning steps in prompt design. Least-to-most \cite{zhou2022least} breaks down reasoning steps into simpler problems. Self-consistency \cite{wang2022self} prompting further enhances LLMs reasoning by ensembling diverse CoT reasoning paths.
Tree-of-Thought (ToT) \cite{yao2023tree} and Graph-of-Thought (GoT) \cite{yao2023beyond, besta2023graph} enable LLMs to explore the thought processes in tree and graph structure, respectively. Moreover, preliminary explorations show that LLMs can use external tools \cite{schick2023toolformer}, be parametric knowledge bases \cite{pan2023unifying}, have theory-of-mind \cite{kosinski2023theory}, act as agents \cite{wang2023survey, xi2023rise}, have graph understanding abilities \cite{wang2023can}, and can serve as optimizers \cite{yang2023large2}.

Apart from using LLMs with frozen parameters, another line of work focuses on efficiently fine-tuning parameters in LLMs, namely parameter efficient fine-tuning, which helps LLMs efficiently adapt to specific tasks, datasets, and domain-specific understanding. Prefix tuning \cite{li2021prefix} keeps the language model parameters frozen and optimizes a small continuous task-specific vector called the prefix. Prompt tuning \cite{lester2021power} is a simpler variant of prefix tuning, where some vectors are prepended at the beginning of a sequence at the input layer. Llama-Adapter \cite{zhang2023llamaadapter} appends a set of learnable prompts as the prefix to the input instruction tokens in the higher transformer layers of Llama \cite{touvron2023llama}.
LoRA \cite{hu2021lora} injects trainable rank decomposition matrices into each layer of the transformer architecture, greatly diminishing the number of trainable parameters for downstream tasks. QLoRA \cite{dettmers2023qlora} updates parameters through a frozen 4-bit LLM into low-rank adapters.


Existing LLMs can be categorized into open-source and API-based models based on accessibility. Open-source models provide access to both model weights and the ability to run the models on local machines, while API-based models restrict users from directly accessing the model weights and only allow them to interact with the models through an API. The open-source LLMs include T5 \cite{raffel2020t5}, Flan-T5 \cite{chung2022scaling}, OPT \cite{zhang2022opt}, BLOOM \cite{scao2022bloom}, GLM \cite{zeng2022glm}, Llama \cite{touvron2023llama}, and Falcon.\footnote{\url{https://falconllm.tii.ae/our-research.html}} By fine-tuning or instruction-tuning Llama, a family of LLMs emerged, such as Alpaca \cite{taori2023alpaca} and Vicuna \cite{chiang2023vicuna}. For API-based models, OpenAI offers four major series of GPT-3 \cite{brown2020language} interface, including \texttt{ada}, \texttt{babbage}, \texttt{curie}, and \texttt{davinci}, corresponding to GPT-3 (350M), GPT-3 (1B), GPT-3 (6.7B), and GPT-3 (175B). GPT-3.5 series includes \text{davinci} and \texttt{turbo}:\texttt{turbo} leverages Reinforcement Learning from Human Feedback (RLHF) \cite{ouyang2022training} to create human-like conversations. GPT-4 \cite{openai2023gpt4} is a well-acknowledged state-of-the-art and achieves astonishing performance on a wide range of tasks. There are some other API-based models such as BARD \cite{manyika2023overview}, Claude,\footnote{\url{https://www.anthropic.com/index/claude-2}} PaLM \cite{chowdhery2022palm}, BloombergGPT \cite{wu2023bloomberggpt}, and LangChain.\footnote{\url{https://www.langchain.com/}}

%

Despite the thriving development of the LLM research, some challenges remain unsolved. For example, LLMs suffer from hallucination issues, that being said, LLMs generate text that is fluent and natural but unfaithful to the source content or under-determined \cite{kaddour2023challenges}. LLMs also have biases due to unfathomable pre-training datasets, including political discourse \cite{feng2023pretraining}, hate speech \cite{huang2023survey}, and discrimination. Moreover, the inference latency of LLMs remains high because of low parallelizability and large memory footprints \cite{pope2023efficiently}. The remaining challenges also include limited context length \cite{kaddour2023challenges}, outdated knowledge \cite{yao2023editing}, misaligned behavior \cite{russell2021human}, brittle evaluation \cite{zhao2021calibrate}, and limited structure understanding ability \cite{chen2023exploring}.



\begin{figure}[t]
    \centering
    \includegraphics[width=0.9\linewidth]{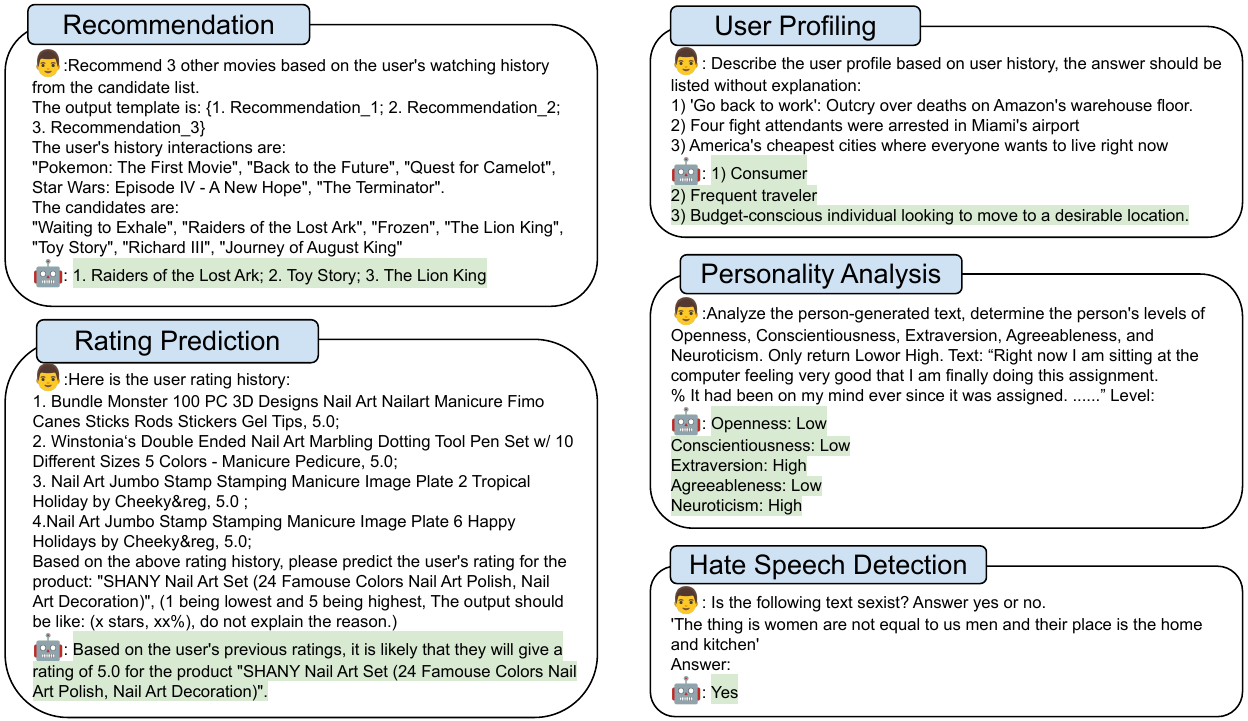}
    \caption{Some examples of LLMs for recommendation, rating prediction, user profiling, personality analysis, and hate speech detection. They serve as compelling examples that demonstrate the ability of LLMs to effectively model, comprehend, and reason based on user-generated content (UGC) and user interactions.}
    \label{fig:motivation}
\end{figure}

\section{Motivation: Why LLMs for User Modeling?}
\label{sec:motivation}
LLMs have demonstrated novel capabilities, exhibiting strong potential in modeling and understanding user-generated content (UGC). In this section, we present related studies and specific examples (Figure \ref{fig:motivation}) to support the claim that LLMs are effective tools for UGC modeling and comprehension.

A growing body of research focuses on utilizing LLMs for recommendation purposes, aiming to predict users' item-based interests from their behavior history. For example, PALR \cite{chen2023palr} generates a user profile based on UGC history and constructs prompts that incorporate the history, profile, and item candidates to enable LLMs to provide recommendations. As illustrated in the real example depicted in Figure \ref{fig:motivation}, LLMs successfully generate reasonable recommendations by considering the user's viewing history. 
In the domain of user profiling, the objective is to summarize users' characteristics, including personality, interests, and topics of interest, based on their generated content and history. Recent research demonstrates that LLMs excel in user profiling. \citet{liu2023once} put behavior history into LLMs and extract users' interest topics and physical regions and thus augment recommender systems. The example in Figure \ref{fig:motivation} further confirms the effectiveness of LLMs in summarizing user characteristics, preferences, and intentions.
In the context of rating prediction, which seeks to comprehend user preferences based on past UGC, \citet{kang2023llms} investigate the ability of LLMs to predict user ratings for candidate items. They find that zero-shot LLMs lag behind traditional recommendation models due to the absence of user interaction data. LLMs can employ reasoning based on users' previous ratings to predict ratings for candidate items.
Recent studies highlight LLMs' capacity to understand user personality. \citet{ji2023chatgpt} employ various prompts to probe LLMs' ability to recognize user personality based on UGC history. LLMs achieve impressive results in personality recognition under zero-shot conditions.
For users with a harmless behavior history, the UM is built upon general interests, content preferences, and interaction styles. However, the UM for users with suspicious behavior history, e.g., hate speech, would have to assess behavior patterns correlated to the suspicious activity, the risk of future incidents, and potentially flag these users for closer monitoring.
From the application aspect, the presence or absence of hate speech in users' history has a significant impact on personalized recommendations. For example, the users with a hate speech history might be steered away from sensitive topics in recommendations. Therefore, suspiciousness detection, e.g., hate speech detection, is an essential application in user modeling.
LLMs are good at detecting suspiciousness in UGC. \citet{del2023respectful} explore zero-shot prompting LLMs for hate speech detection. They find that zero-shot prompting can achieve performance comparable to and even surpass fine-tuned models. Figure \ref{fig:motivation} presents an example to further validate LLMs' effectiveness in detecting suspicious UGC.

Collectively, these studies and examples demonstrate LLMs' capabilities of modeling, understanding, and reasoning UGC and user behavior. They provide comprehensive evidence that LLMs can serve as valuable tools for user modeling, showcasing significant potential to improve user-oriented applications.



\section{Taxonomies}
\label{sec:taxonomy}

In this section, we present taxonomies that classify LLM-UM techniques based on their approaches and applications.
In Figure \ref{fig:overview}, we structure the survey according to these two taxonomies in Section \ref{sec:approach} and \ref{sec:application}, where the upper-level section is the parent category and the lower-level section is the child concept. For instance, the concept of ``Approaches" to LLM-UM encompasses ``LLMs-as-Predictors", ``LLMs-as-Enhancers", ``LLMs-as-Controllers", and ``LLMs-as-Evaluators".

\subsection{Approach Perspective}
\label{sec:taxo_approach}
LLMs play diverse roles in LLM-UM systems. Based on their functionality, LLM-UM work can be categorized into four distinct approaches.
(1) \textbf{LLMs-as-Predictors} involves leveraging LLMs for reasoning and generating answers directly. Depending on the role of LLMs, these LLM-UM methods can further be categorized as common generative reasoners for complex tasks, agents/simulators that model and predict human behavior, classifiers/detectors, scoring functions, explanation generation, and Chatbots for user modeling. (2) \textbf{LLMs-as-Enhancers} refers to using LLMs as augmentation modules to enhance the downstream user modeling system. LLMs can act as profilers to infer user preferences and characteristics, serve as feature encoders to generate latent UGC representations, augment discriminative user modeling systems with knowledge stored in LLMs, and generate high-quality data for small UM model training. (3) LLMs also possess the ability to control the pipeline of UM systems (\textbf{LLMs-as-Controllers}), automatically determining whether to execute certain operations. (4) LLMs can also serve as evaluators (\textbf{LLMs-as-Evaluators}) to score and analyze conversations and text generated under open-domain settings.

\subsection{Application Perspective}
\label{sec:taxo_application}
Another taxonomy for categorizing LLM-UM is based on the downstream applications they address. Generally, LLM-UM systems aim to adapt to users' personal needs and detect suspiciousness in user data.

\textbf{Personalization} refers to tailoring experiences to individual preferences. Existing LLM-UM works can be categorized as follows: user profiling, which aims to tell the characteristics, personalities, stances, and sentiments towards certain targets; personalized recommendation gives the recommended items based on the user's behavior history and user profile; personalized assistants aim to adapt to users' specific needs and provide customized assistance; personalized dialog systems that interact iteratively with humans based on user data; and personalized applications in the education and healthcare domains. \textbf{Suspiciousness detection} is focused on identifying malicious users and UGC such as fraudsters, spammers, discriminations, and misinformation to preserve the integrity of social discourse \cite{jiang2016suspicious, tan2023botpercent}. Existing LLM-UM works in suspiciousness detection can be categorized as follows: fraud detection for identifying malicious users or information in social platforms; discrimination detection to combat the spread of hate speech, predatory, and sexist; misinformation detection to identify fake news and propaganda; and LLM-generated text detection to identify if text is AI-generated for the integrity of education, online discourse, etc.



\section{Approaches to LLM-UM}

\label{sec:approach}
Given the strong capabilities in generation \cite{zhao2023survey}, reasoning \cite{wei2022chain}, knowledge comprehension \cite{sun2023head}, and a good understanding of UGC as elaborated in Section \ref{sec:motivation}, LLMs can be used to supercharge UM systems. The LLM-UM approaches can be generally categorized into three categories based on the LLMs' role, 
where the first envisions LLMs as the sole predictor that generates prediction directly, 
the second employs LLMs as enhancers to probe more information for the UM system augmentation, 
the third empowers LLMs with the ability to control the UM methods pipeline, automating the UM process, and the last uses LLMs as evaluators, assessing the performance of the system.
It's worth mentioning that the form of ``user model" in LLM-UM remains consistent with the previous definition, encompassing the knowledge and patterns that are discovered with the help of user-generated content, and user-user/item interaction networks \cite{he2023survey}. The distinction of LLM-UM from previous paradigms lies in the approaches, where LLM-UM are empowered or enhanced by LLMs to gain user-related knowledge.
In the following subsections, we summarize each paradigm and present representative approaches.

\subsection{LLMs-as-Predictors}
\label{sec:approach_predictor}
In this section, we introduce the LLMs-as-Predictors paradigm presented in LLM-UM works, which means LLMs are leveraged to make predictions and generate answers for downstream applications directly. More specifically, approaches in leveraging LLMs as generative reasoners, simulators/agents, classifiers/detectors, scoring/ranking functions, explainers, and Chatbots.

\begin{figure}[t]
    \centering
    \includegraphics[width=0.98\linewidth]{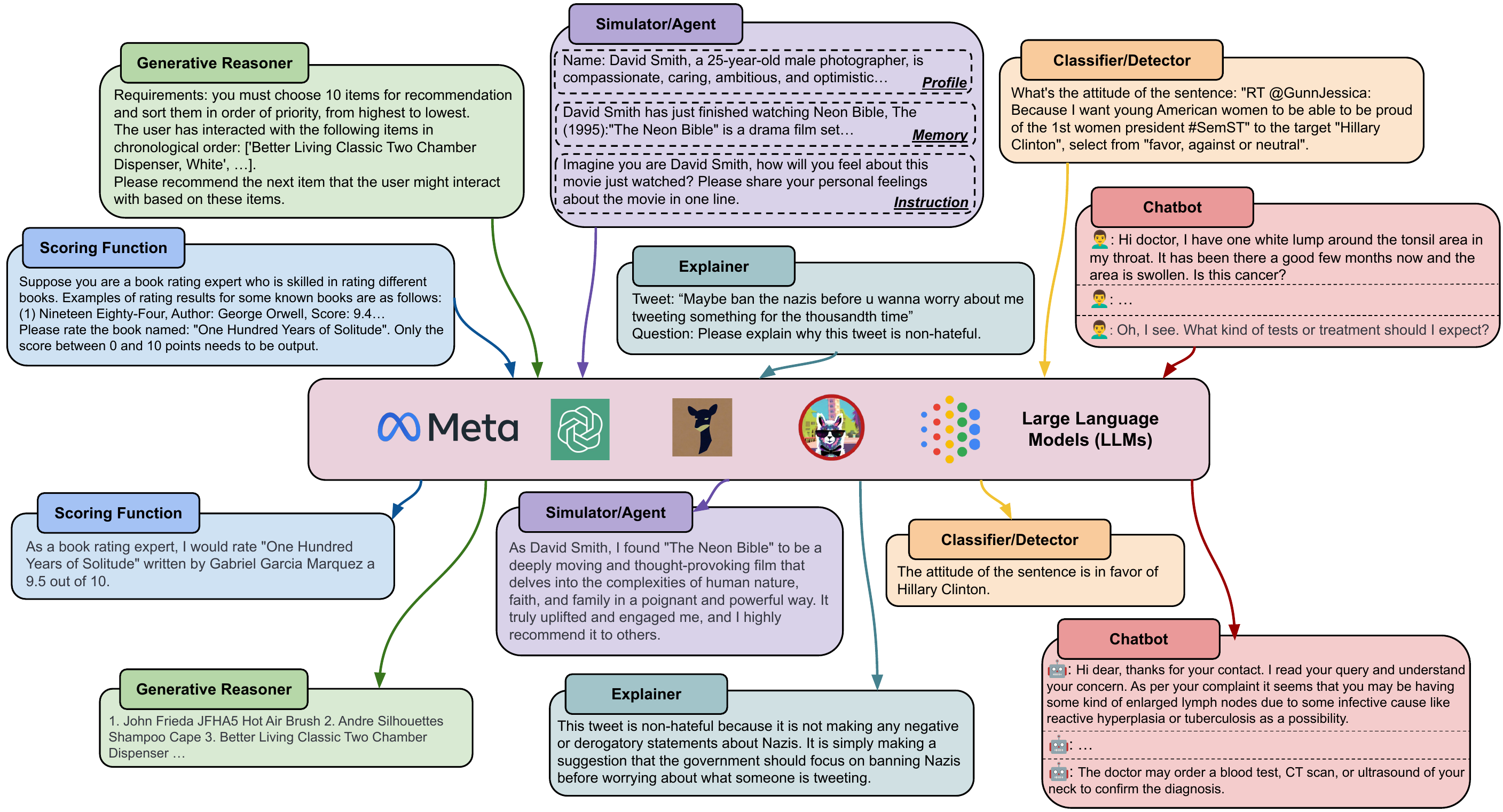}
    \caption{LLMs-as-Predictors, where LLMs are exclusively utilized to generate the predicted response.}
    \label{fig:enter-label}
\end{figure}

\subsubsection{Common Generative Reasoner}
Thanks to the strong generalization of LLMs, numerous UM systems incorporate them as generative reasoners to directly execute downstream tasks with a method design that is free of shenanigans. Considering the significant body of work falling under the generative reasoner category, we further classify them into non-tuning and fine-tuning methods based on the parameter status of LLMs. 

\paragraph{Non-tuned LLMs}
Utilizing LLMs as non-fine-tuning generative reasoners involves directly employing pretrained LLMs without adjusting their parameters for specific tasks. This approach preserves their versatility and ability to generalize across diverse circumstances. Therefore, the focus of non-tune LLMs research is mainly on prompt template and pipeline design.
\citet{di2023evaluating} present a comprehensive experimental evaluation framework to examine LLMs' ability in recommendation rigorously. \citet{liu2023chatgpt} investigate LLMs capability in recommendation by designing prompts to test ChatGPT on directly performing rating prediction, sequential recommendation, direct recommendation, explanation, generation, and review summarization. NIR~\cite{wang2023zero} designs a three-stage prompt template that guides LLMs to understand user preference, select representative movie-watching history, and recommend a list of movies. \citet{sanner2023large} explore the prompting LLMs with item-based and language-based preference and find LLMs are strong near cold-start recommendation systems for pure language-based preference. BookGPT \cite{zhiyuli2023bookgpt} design prompt templates and the overall pipeline to perform book rating recommendation, user rating recommendation, and book summary recommendation. \citet{hou2023large} design prompting template by including the sequential interaction history, candidate items, and ranking instruction and show that LLMs have promising zero-shot ranking abilities in recommendation tasks. PALR~\cite{chen2023palr} designs prompt templates containing user interaction history sequences, LLM-generated user profiles, and candidates, then feeds them into a general-purpose LLM for recommendation. LaMP~\cite{salemi2023lamp} offers an evaluation framework for personalized content generation and classification. \citet{li2023teach} design a multistage prompting strategy and multitasking to help frozen LLMs generate personalized content. The multistage contains retrieval, ranking, summarization, synthesis, and generation, and multitasking is to predict if the same author writes two documents. \citet{li2023preliminary} investigate ChatGPT's capabilities in personalized news recommendation, news provider fairness, and fake news detection, where they find ChatGPT is sensitive to input phrasing. \citet{rao2023can} explore LLMs capabilities to analyze human personalities based on the Myers-Briggs Type Indicator (MBTI) test. \citet{ji2023chatgpt} investigate the text-based personality recognition ability of ChatGPT and propose a level-oriented prompting strategy to optimize zero-shot chain-of-thought performance on personality recognition. \citet{ghanadian2023chatgpt} leverage ChatGPT to assess the suicide risks on social media with zero-shot and few-shot prompting. \citet{fan2023uncovering} investigate ChatGPT's capabilities in discourse dialogue analysis, including topic segmentation, discourse relation recognition, and discourse parsing. \citet{wu2023large} prompt LLMs to pairwise compare lawmakers and then scale the resulting graph using the Bradley-Terry model, finding that LLMs can be used to estimate the latent positions of politicians.

\paragraph{Fine-tuned LLMs}
Fine-tuning LLMs can facilitate their adaptation to specific UM tasks and lead to better domain specialization. Given the massive number of parameters, research generally employs parameter-efficient fine-tuning (PEFT) techniques, such as LoRA, Llama-adapter, prompt-tuning, etc. InstructRec \cite{zhang2023recommendation} considers LLMs as generative instruction following recommender and performs instruction tuning on Flan-T5-XL \cite{chung2022scaling} model with a large amount of user-personalized instruction data. GIRL \cite{zheng2023generative} proposes a Proximal Policy Optimization (PPO)-based reinforcement learning method to fine-tune LLMs, which aims to improve the LLM's ability to assess the compatibility between a job and a user. TallRec \cite{bao2023tallrec} leverages rec tuning samples containing user history behavior, new items, and feedback for instruction tuning, and it uses LoRA to improve the efficiency as well. GLRec \cite{wu2023exploring} constructs prompts based on meta paths extracted from the user behavior graph, then leverages weighted path embeddings, instruction tuning, and LoRA for LLMs tuning and generates recommended items. \citet{tie2023automatic} leverage fine-tuned LLMs to provide clinically useful, personalized impressions.

Combining non-tune and fine-tune paradigms can improve LLM performance. For example, LaMP \cite{salemi2023lamp} proposes a retrieval-augmented approach to retrieve personalized history to construct prompts for LLM generation under zero-shot and fine-tuning settings. \citet{christakopoulou2023large} investigate the user's interest journey using few-shot prompting, prompt-tuning, and fine-tuning for journey name extraction. ReLLa \cite{lin2023rella} proposes to retrieve semantic user behavior to augment LLMs under the zero-shot settings and design retrieval-enhanced instruction tuning for the few-shot recommendation.

\subsubsection{Simulator/Agent}
Using LLMs as autonomous agents has been a prosperous research direction recently, which expects LLMs to accomplish tasks through self-directed planning and actions \cite{wang2023survey}. In the user modeling domain, LLMs can serve as a user simulator to imitate user behavior, conduct UM applications with planning and actions, use external tools, etc. Some works leverage LLM as a user simulator to predict user behavior. RecLLM \cite{friedman2023leveraging} plugs LLM into the conversational recommender system to generate synthetic conversations to simulate user behavior for tuning system modules. UGRO \cite{hu2023unlocking} uses LLM as an annotation-free user simulator to assess dialogue responses.
\citet{kong2023large} fine-tune LLMs on genuine human-machine conversations to get a user simulator to generate a high-quality human-centric synthetic conversation dataset. PersonaLLM \cite{jiang2023personallm} investigates whether the behavior of LLM-generated personas can reflect certain personality traits accurately and consistently.

A few research studies have envisioned LLMs as agents and enabled them to interact with and explore the environment to gain a better understanding of user modeling tasks.
Generative Agent \cite{park2023generative} leverages LLMs to simulate human behavior in a social context with a memory module to record agent history and memories, and retrieve them for planning.
RecMind \cite{wang2023recmind} presents an LLM-powered autonomous recommender agent capable of providing precise, personalized recommendations through careful planning by utilizing tools for obtaining external knowledge and leveraging individual data. InteRecAgent \cite{huang2023recommender} employs LLMs as the brain and recommender models as tools. It comprises key components such as a memory bus, dynamic demonstration-augmented task planning, and reflection. RecAgent \cite{wang2023recagent} constructs a user simulator that regards each user as an LLM-based autonomous agent and lets different agents freely communicate, behave, and evolve within the recommender system. AutoGPT\footnote{\url{https://news.agpt.co/}} demonstrates autonomous comprehension of specific objectives using natural language and carries out automated processes in a continuous loop, effectively accomplishing user-specific tasks. LLMs as agents can also empower LLMs with external tools and API for better user understanding. For example, Graph-Toolformer \cite{zhang2023graph} teaches LLMs to use external graph-related API to augment LLMs' ability to reason over structural data and, therefore perform sequential recommendation and user rating prediction.


\begin{table}[!t] 
\centering
\caption{Representative approaches using the LLMs-as-Predictors paradigm.}
\label{tab:approach}
\renewcommand\arraystretch{1.25}
\resizebox{0.9\linewidth}{!}{
\begin{tabular}{cccc}
\toprule[1.5pt]
\multicolumn{1}{c|}{\makecell[c]{\textbf{Roles}}} & \multicolumn{1}{c|}{\makecell[c]{\textbf{Applications}}} & \multicolumn{1}{c|}{\makecell[c]{\textbf{LLM Backbones}}} & \multicolumn{1}{c}{\makecell[c]{\textbf{Models}}}\\ \bottomrule

\multicolumn{1}{c|}{\multirow{16}{*}{\makecell[c]{\\ \\ Common Generative Reasoner}}} & \multicolumn{1}{c|}{\multirow{4}{*}{\makecell[c]{Recommendation}}} & \multicolumn{1}{c|}{GPT family} &
\begin{tabular}[c]{@{}c@{}}
\citet{liu2023chatgpt}, NIR \cite{wang2023zero}, BookGPT \cite{zhiyuli2023bookgpt},\\ \citet{hou2023large}, \citet{li2023preliminary}, \citet{di2023evaluating}
\end{tabular} \\ \cline{3-4}

\multicolumn{1}{c|}{} & \multicolumn{1}{c|}{} & \multicolumn{1}{c|}{PaLM} &
\begin{tabular}[c]{@{}c@{}}
\citet{sanner2023large}
\end{tabular} \\ \cline{3-4}

\multicolumn{1}{c|}{} & \multicolumn{1}{c|}{} & \multicolumn{1}{c|}{Llama/Vicuna} &
\begin{tabular}[c]{@{}c@{}}
PALR \cite{chen2023palr}, GIRL \cite{zheng2023generative}, TallRec \cite{bao2023tallrec}, GLRec \cite{wu2023exploring}, ReLLa \cite{lin2023rella}
\end{tabular} \\ \cline{3-4}

\multicolumn{1}{c|}{} & \multicolumn{1}{c|}{} &  \multicolumn{1}{c|}{FLAN-T5} &
\begin{tabular}[c]{@{}c@{}}
\citet{zhang2023recommendation}
\end{tabular} \\ \cline{2-4}

\multicolumn{1}{c|}{} & \multicolumn{1}{c|}{\multirow{3}{*}{\makecell[c]{Intelligent Assistant}}} & \multicolumn{1}{c|}{GPT family} &
\begin{tabular}[c]{@{}c@{}}
LaMP \cite{salemi2023lamp}, \citet{chakrabarty2023creativity}
\end{tabular} \\ \cline{3-4}

\multicolumn{1}{c|}{} & \multicolumn{1}{c|}{}  & \multicolumn{1}{c|}{FLAN-T5/T5} &
\begin{tabular}[c]{@{}c@{}}
LaMP \cite{salemi2023lamp}, \citet{li2023teach}
\end{tabular} \\ \cline{3-4}

\multicolumn{1}{c|}{} & \multicolumn{1}{c|}{}  & \multicolumn{1}{c|}{ChatGLM} &
\begin{tabular}[c]{@{}c@{}}
Disc-LawLLM \cite{yue2023disc}
\end{tabular} \\ \cline{2-4}

\multicolumn{1}{c|}{} & \multicolumn{1}{c|}{\multirow{2}{*}{\makecell[c]{User Profiling}}}  & \multicolumn{1}{c|}{ChatGPT} &
\begin{tabular}[c]{@{}c@{}}
\citet{rao2023can}, \citet{ji2023chatgpt}, \citet{wu2023large}
\end{tabular} \\ \cline{3-4}

\multicolumn{1}{c|}{} & \multicolumn{1}{c|}{}  & \multicolumn{1}{c|}{LaMDA, PaLM} &
\begin{tabular}[c]{@{}c@{}}
\citet{christakopoulou2023large}
\end{tabular} \\ \cline{2-4}

\multicolumn{1}{c|}{} & \multicolumn{1}{c|}{\multirow{1}{*}{\makecell[c]{Dialogue System}}}  & \multicolumn{1}{c|}{ChatGPT} &
\begin{tabular}[c]{@{}c@{}}
\citet{fan2023uncovering}
\end{tabular} \\ \cline{2-4}

\multicolumn{1}{c|}{} & \multicolumn{1}{c|}{\multirow{2}{*}{\makecell[c]{Education}}}  & \multicolumn{1}{c|}{ChatGPT} &
\begin{tabular}[c]{@{}c@{}}
\citet{sharma2023performance}, \citet{elkins2023useful}, \\ \citet{ochieng2023large}, C-LLM \cite{olga2023generative}, \citet{phung2023generative}
\end{tabular} \\ \cline{3-4}

\multicolumn{1}{c|}{} & \multicolumn{1}{c|}{}  & \multicolumn{1}{c|}{BARD} &
\begin{tabular}[c]{@{}c@{}}
\citet{ochieng2023large}
\end{tabular} \\ \cline{2-4}

\multicolumn{1}{c|}{} & \multicolumn{1}{c|}{\multirow{4}{*}{\makecell[c]{Healthcare}}}  & \multicolumn{1}{c|}{GPT family} &
\begin{tabular}[c]{@{}c@{}} 
\citet{wang2023large}, \citet{ghanadian2023chatgpt}, \\ \citet{tie2023automatic}, PharmacyGPT \cite{liu2023pharmacygpt}, \citet{zhang2023potential}, \\ \citet{fu2023enhancing}, Mental-LLM \cite{xu2023mental}, \citet{peters2023large}
\end{tabular} \\ \cline{3-4}

\multicolumn{1}{c|}{} & \multicolumn{1}{c|}{}  & \multicolumn{1}{c|}{PaLM} &
\begin{tabular}[c]{@{}c@{}}
\citet{liu2023large}
\end{tabular} \\ \cline{3-4}

\multicolumn{1}{c|}{} & \multicolumn{1}{c|}{}  & \multicolumn{1}{c|}{FLAN-T5} &
\begin{tabular}[c]{@{}c@{}}
Mental-LLM \cite{xu2023mental}
\end{tabular} \\ \cline{3-4}

\multicolumn{1}{c|}{} & \multicolumn{1}{c|}{}  & \multicolumn{1}{c|}{Llama/Alpaca/Vicuna} &
\begin{tabular}[c]{@{}c@{}}
\citet{tie2023automatic}, Mental-LLM \cite{xu2023mental}
\end{tabular} \\ \cline{1 -4}

\multicolumn{1}{c|}{\multirow{5}{*}{\makecell[c]{Simulator/Agent}}} & \multicolumn{1}{c|}{\multirow{2}{*}{\makecell[c]{Recommendation}}}  & \multicolumn{1}{c|}{GPT family} &
\begin{tabular}[c]{@{}c@{}} 
RecMind \cite{wang2023recmind}, InteRecAgent \cite{huang2023recommender}, \\ RecAgent \cite{wang2023recagent}, Graph-Toolformer \cite{zhang2023graph}
\end{tabular} \\ \cline{3-4}

\multicolumn{1}{c|}{} & \multicolumn{1}{c|}{\multirow{1}{*}{\makecell[c]{}}}  & \multicolumn{1}{c|}{LaMDA} &
\begin{tabular}[c]{@{}c@{}}
RecLLM \cite{friedman2023leveraging}
\end{tabular} \\ \cline{2-4}

\multicolumn{1}{c|}{} & \multicolumn{1}{c|}{\multirow{2}{*}{\makecell[c]{Dialogue System}}}  & \multicolumn{1}{c|}{ChatGPT} &
\begin{tabular}[c]{@{}c@{}}
UGRO \cite{hu2023unlocking}
\end{tabular} \\ \cline{3-4}

\multicolumn{1}{c|}{} & \multicolumn{1}{c|}{} & \multicolumn{1}{c|}{Llama} &
\begin{tabular}[c]{@{}c@{}}
\citet{kong2023large}
\end{tabular} \\ \cline{2-4}

\multicolumn{1}{c|}{} & \multicolumn{1}{c|}{Intelligent Assistant} & \multicolumn{1}{c|}{GPT-3.5} &
\begin{tabular}[c]{@{}c@{}}
PersonaLLM \cite{jiang2023personallm}
\end{tabular} \\ \cline{1-4}

\multicolumn{1}{c|}{\multirow{11}{*}{\makecell[c]{Classifier/Detector}}} & \multicolumn{1}{c|}{\multirow{2}{*}{\makecell[c]{User Profiling}}}  & \multicolumn{1}{c|}{GPT family} &
\begin{tabular}[c]{@{}c@{}}
\citet{zhang2022would}, LoT \cite{hu2023ladder}, \citet{mu2023navigating},\\ SentimentGPT \cite{kheiri2023sentimentgpt}, \citet{ziems2023can}
\end{tabular} \\ \cline{3-4}

\multicolumn{1}{c|}{} & \multicolumn{1}{c|}{} & \multicolumn{1}{c|}{LLaMA} &
\begin{tabular}[c]{@{}c@{}}
\citet{mu2023navigating}
\end{tabular} \\ \cline{2-4}

\multicolumn{1}{c|}{} & \multicolumn{1}{c|}{\multirow{1}{*}{\makecell[c]{Healthcare}}}  & \multicolumn{1}{c|}{\multirow{1}{*}{\makecell[c]{GPT family}}} &
\begin{tabular}[c]{@{}c@{}}
\citet{qi2023evaluating}, \citet{qin2023read}, ALEX \cite{jiang2023balanced} 
\end{tabular} \\ \cline{2-4}

\multicolumn{1}{c|}{} & \multicolumn{1}{c|}{\multirow{2}{*}{\makecell[c]{Fraud Detection}}}  & \multicolumn{1}{c|}{\multirow{1}{*}{\makecell[c]{GPT family}}} &
\begin{tabular}[c]{@{}c@{}}
\citet{shukla2023catch}, \citet{zhu2023clickbait}
\end{tabular} \\ \cline{3-4}

\multicolumn{1}{c|}{} & \multicolumn{1}{c|}{}  & \multicolumn{1}{c|}{\multirow{1}{*}{\makecell[c]{T5}}} &
\begin{tabular}[c]{@{}c@{}}
Spam-T5 \cite{labonne2023spam}
\end{tabular} \\ \cline{2-4}

\multicolumn{1}{c|}{} & \multicolumn{1}{c|}{\multirow{3}{*}{\makecell[c]{Discrimination Detection}}}  & \multicolumn{1}{c|}{\multirow{1}{*}{\makecell[c]{GPT-3}}} &
\begin{tabular}[c]{@{}c@{}}
\citet{chiu2021detecting}
\end{tabular} \\ \cline{3-4}

\multicolumn{1}{c|}{} & \multicolumn{1}{c|}{}  & \multicolumn{1}{c|}{\multirow{1}{*}{\makecell[c]{Llama}}} &
\begin{tabular}[c]{@{}c@{}}
\citet{nguyen2023fine}
\end{tabular} \\ \cline{3-4}

\multicolumn{1}{c|}{} & \multicolumn{1}{c|}{}  & \multicolumn{1}{c|}{\multirow{1}{*}{\makecell[c]{FLAN-T5}}} &
\begin{tabular}[c]{@{}c@{}}
\citet{del2023respectful}
\end{tabular} \\ \cline{2-4}

\multicolumn{1}{c|}{} & \multicolumn{1}{c|}{\multirow{2}{*}{\makecell[c]{Misinformation Detection}}}  & \multicolumn{1}{c|}{\multirow{1}{*}{\makecell[c]{ChatGPT}}} &
\begin{tabular}[c]{@{}c@{}}
\citet{li2023preliminary}
\end{tabular} \\ \cline{3-4}

\multicolumn{1}{c|}{} & \multicolumn{1}{c|}{}  & \multicolumn{1}{c|}{\multirow{1}{*}{\makecell[c]{Llama}}} &
\begin{tabular}[c]{@{}c@{}}
\citet{pavlyshenko2023analysis}
\end{tabular} \\ \cline{2-4}

\multicolumn{1}{c|}{} & \multicolumn{1}{c|}{\multirow{1}{*}{\makecell[c]{LLM-Gen Text Detection}}}  & \multicolumn{1}{c|}{\multirow{1}{*}{\makecell[c]{ChatGPT}}} &
\begin{tabular}[c]{@{}c@{}}
\citet{bhattacharjee2023fighting}
\end{tabular} \\ \cline{1-4}


\multicolumn{1}{c|}{\multirow{5}{*}{\makecell[c]{Scoring Function}}} & \multicolumn{1}{c|}{\multirow{3}{*}{\makecell[c]{Recommendation}}}  & \multicolumn{1}{c|}{ChatGPT} &
\begin{tabular}[c]{@{}c@{}}
Chat-REC \cite{gao2023chat}, \citet{liu2023chatgpt}, \citet{kang2023llms},\\ BookGPT \cite{zhiyuli2023bookgpt}, \citet{dai2023uncovering}
\end{tabular} \\ \cline{3-4}

\multicolumn{1}{c|}{} & \multicolumn{1}{c|}{} & \multicolumn{1}{c|}{LaMDA} &
\begin{tabular}[c]{@{}c@{}}
RecLLM \cite{friedman2023leveraging}
\end{tabular} \\ \cline{3-4}

\multicolumn{1}{c|}{} & \multicolumn{1}{c|}{}  & \multicolumn{1}{c|}{Llama} &
\begin{tabular}[c]{@{}c@{}}
TallRec \cite{bao2023tallrec}
\end{tabular} \\ \cline{2-4}

\multicolumn{1}{c|}{} & \multicolumn{1}{c|}{\multirow{1}{*}{\makecell[c]{Dialogue System}}}  & \multicolumn{1}{c|}{\multirow{1}{*}{\makecell[c]{ChatGPT}}} &
\begin{tabular}[c]{@{}c@{}}
\citet{hu2023unlocking}
\end{tabular} \\ \cline{2-4}

\multicolumn{1}{c|}{} & \multicolumn{1}{c|}{\multirow{1}{*}{\makecell[c]{Misinformation Detection}}} & \multicolumn{1}{c|}{ChatGPT} &
\begin{tabular}[c]{@{}c@{}}
\citet{yang2023large}
\end{tabular} \\ \cline{1-4}

\multicolumn{1}{c|}{\multirow{4}{*}{\makecell[c]{Explainer}}} & \multicolumn{1}{c|}{\multirow{1}{*}{\makecell[c]{Recommendation}}}  & \multicolumn{1}{c|}{ChatGPT} &
\begin{tabular}[c]{@{}c@{}}
Chat-REC \cite{gao2023chat}, \citet{liu2023chatgpt}, 
\end{tabular} \\ \cline{2-4}

\multicolumn{1}{c|}{} &\multicolumn{1}{c|}{\multirow{1}{*}{\makecell[c]{Discrimination Detection}}}  & \multicolumn{1}{c|}{GPT family} &
\begin{tabular}[c]{@{}c@{}}
\citet{wang2023evaluating}, \citet{ziems2023can}
\end{tabular} \\ \cline{2-4}

\multicolumn{1}{c|}{} & \multicolumn{1}{c|}{Education}  & \multicolumn{1}{c|}{Llama/Vicuna} &
\begin{tabular}[c]{@{}c@{}}
\citet{bao2023exploring}
\end{tabular} \\ \cline{2-4}

\multicolumn{1}{c|}{} & \multicolumn{1}{c|}{Healthcare}  & \multicolumn{1}{c|}{GPT-3.5} &
\begin{tabular}[c]{@{}c@{}}
ALEX \cite{jiang2023balanced}
\end{tabular} \\ \cline{1-4}

\multicolumn{1}{c|}{\multirow{6}{*}{\makecell[c]{Chatbot}}} & \multicolumn{1}{c|}{\multirow{1}{*}{\makecell[c]{Recommendation}}}  & \multicolumn{1}{c|}{ChatGPT} &
\begin{tabular}[c]{@{}c@{}}
Chat-REC \cite{gao2023chat}, \citet{lin2023sparks}, \citet{he2023large}, GeneRec \cite{wang2023generative}, 
\end{tabular} \\ \cline{2-4}

\multicolumn{1}{c|}{} & \multicolumn{1}{c|}{\multirow{1}{*}{\makecell[c]{Dialogue System}}} & \multicolumn{1}{c|}{GPT family} &
\begin{tabular}[c]{@{}c@{}}
DiagGPT \cite{cao2023diaggpt}, \citet{cho2022personalized}, RefGPT \cite{yang2023refgpt}, \\ \citet{hudevcek2023llms}, \citet{tu2023characterchat}, \citet{zheng2023building}
\end{tabular} \\ \cline{2-4}


\multicolumn{1}{c|}{} & \multicolumn{1}{c|}{\multirow{1}{*}{\makecell[c]{Intelligent Assistant}}}  & \multicolumn{1}{c|}{\multirow{1}{*}{\makecell[c]{ChaGPT}}} &
\begin{tabular}[c]{@{}c@{}}
\citet{lakkaraju2023can}, \citet{hassan2023chatgpt}
\end{tabular} \\ \cline{2-4}

\multicolumn{1}{c|}{} & \multicolumn{1}{c|}{\multirow{1}{*}{\makecell[c]{Healthcare}}}  & \multicolumn{1}{c|}{\multirow{1}{*}{\makecell[c]{ChaGPT}}} &
\begin{tabular}[c]{@{}c@{}}
\citet{chen2023llm}, ChatDoctor \cite{li2023chatdoctor}
\end{tabular} \\ \cline{2-4}

\multicolumn{1}{c|}{} & \multicolumn{1}{c|}{Education} & \multicolumn{1}{c|}{Llama} &
\begin{tabular}[c]{@{}c@{}}
EduChat \cite{dan2023educhat}
\end{tabular} \\

\bottomrule[1.5pt]
\end{tabular}
}

\end{table}

\subsubsection{Classifier/Detector}
LLMs can be prompted to serve as classifiers and detectors to analyze UGC, e.g., detect stance, hate speech, and suspicious behavior \cite{robinson2022leveraging, sun2023text}. 
\citet{zhang2022would} discuss the potential of LLMs in stance detection and explanation generation. \citet{hu2023ladder} propose ladder-of-thought (LoT) that assimilates high-quality external knowledge in small LMs to augment stance detection in LLMs.
\citet{mu2023navigating} investigate LLMs potential in zero-shot text classification in computational social science, including bragging, vaccine, complaint, and hate speech detection. \citet{ziems2023can} comprehensively investigate LLMs capabilities in classifying dialect features, emotion, humor, empathy, discourse acts, and more detection and classification tasks. \citet{parikh2023exploring} explores LLMs capabilities in user intent classification under zero-shot and few-shot settings. 
\citet{qin2023read} construct an interactive depression detection system based on LLMs with CoT and a selector of users' tweets. \citet{ferrara2023social} discuss social bot detection in the era of LLMs, which points out that LLMs can be used to improve bot detection in low-resource language and domain. \citet{qi2023evaluating} design zero-shot, few-shot, and fine-tuning methods to perform suicidal risk and cognitive distortion identification on Chinese social media. SentimentGPT \cite{kheiri2023sentimentgpt} investigates the GPT's sentiment analysis capabilities in prompt engineering and fine-tuning. ALEX \cite{jiang2023balanced} prompts an LLM with the predicted results from BERT models for health data classification. \citet{finch2023leveraging} leverage GPT-3.5 to perform dialogue behavior detection for nine categories. \citet{bhattacharjee2023fighting} use ChatGPT to detect LLM-generated text. There are also works focused on designing prompts to enable LLMs to detect suspiciousness content such as clickbait \cite{zhu2023clickbait}, hate speech \cite{chiu2021detecting}, and sexual predatory \cite{nguyen2023fine}. 

\subsubsection{Scoring Function}
LLMs can also be used as scoring or ranking functions that rate and rank items based on the user's behavior history. Chat-REC \cite{gao2023chat} and \citet{liu2023chatgpt} leverage frozen LLMs to conduct item rating prediction. \citet{kang2023llms} comprehensively evaluate LLMs item rating prediction in zero-shot, few-shot, and fine-tuning scenarios. \citet{dai2023uncovering} formulate point-wise, pair-wise, and list-wise to elicit LLM's item scoring ability. BookGPT \cite{zhiyuli2023bookgpt} employs LLMs to predict both book ratings and personalized user ratings on books. TallRec \cite{bao2023tallrec} employs instruction tuning for book and movie rating prediction. RecLLM \cite{friedman2023leveraging} takes LLMs as a ranking and explanation module by feeding side information to produce a score for an item and an explanation for the score. \citet{hu2023unlocking} use LLMs to predict the user's satisfaction score with a conversation response and select the response with the highest satisfaction as output.

\subsubsection{Explainer}
The strong generation and reasoning ability make LLMs good explainers for the UM system, making its process understandable to humans. 
\citet{liu2023chatgpt} prompts LLMs to generate recommendation explanations based on the user's interaction history. Chat-REC \cite{gao2023chat} explains the recommendation in the interaction process with users. \citet{yu2023temporal} fine-tune Open-Llama to generate an explanation of the financial time series data forest. ALEX \cite{jiang2023balanced} uses LLM to generate an explanation for health-related UGC classification and would self-correct the prediction after explaining. \citet{bao2023exploring} propose a self-reinforcement LLM-based framework in learnersouring to generate student-aligned explanations, evaluate, and iteratively enhance explanations automatically. \citet{wang2023evaluating} use LLMs to generate explanations for hateful and non-hateful content. \citet{ziems2023can} investigate LLMs capabilities in social content explanation generation tasks like figurative language explanation and implied misinformation explanation.

\subsubsection{Chatbot}
LLMs are often used as chatbots to empower the UM system's interactivity and enhance understanding of UGC.
Chat-REC \cite{gao2023chat} employs ChatGPT as a ChatBot to interact with human by incorporating user queries with user profiles and user behavior history. \citet{lin2023sparks} observe ChatGPT's behavior in recommendation-oriented dialogues and demonstrate the potential for ChatGPT to serve as an artificial general recommender (AGR). \cite{qin2023read} use ChatGPT as a chat interface with humans to understand the humans' mental status based on their tweets and responses. \citet{lakkaraju2023can} investigate LLMs' potential in serving as personal financial advisors, where LLMs iteratively interact with humans as a chatbot. \citet{hassan2023chatgpt} present a framework that enables LLMs to act as personal data scientists, where users interact with an LLM-based chatbot, and LLMs would return data analysis. CharacterChat \cite{tu2023characterchat} designs prompts with behavior preset and dynamic memory to help LLMs act as a chatbot with a specific personality. \citet{zheng2023building} augment Llama with ChatGPT-generated dataset for optimized emotional support chatbot. \citet{he2023large} empirically study conversational recommendation with LLMs by constructing in-the-wild datasets. ChatDoctor \cite{li2023chatdoctor} fine-tunes Llama using 100,000 patient-doctor dialogues to create a specialized Chatbot with enhanced accuracy in medical advice. \citet{chen2023llm} propose a three-stage pipeline to design prompts for ChatGPT to serve as both doctor and patient chatbots. GeneRec \cite{wang2023generative} takes ChatGPT as a user conversational interface and takes user instruction and feedback to generate personalized content.

\subsection{LLMs-as-Enhancers}
In this section, we analyze the approaches that leverage LLMs as enhancers in the UM models. That being said, LLMs are not used to generate task answers directly but are leveraged to serve as plug-in augmentation modules instead. Approaches use LLMs as profilers, feature encoders, knowledge augmenters, and data generators.

\begin{figure}[t]
    \centering
    \includegraphics[width=0.95\linewidth]{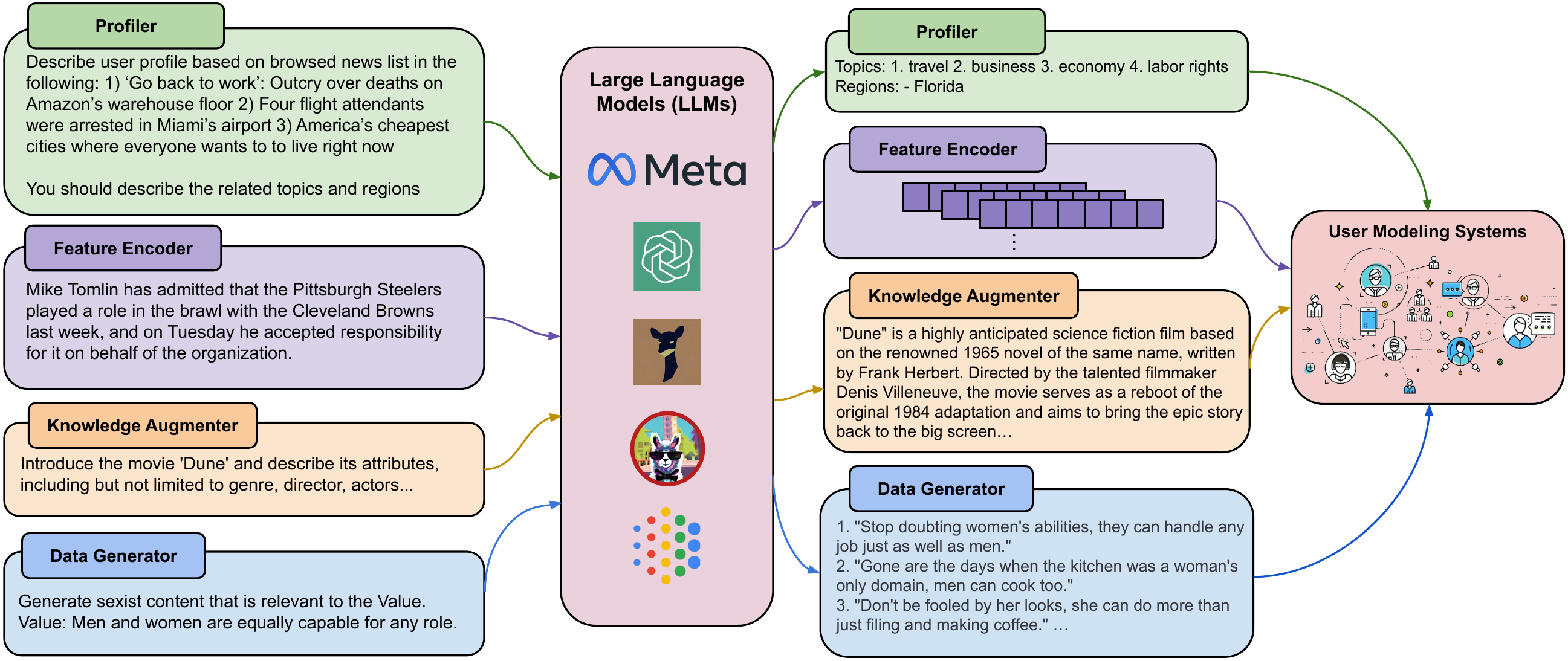}
    \caption{LLMs-as-Enhancers, where LLMs are leveraged to generate user profiles, content embeddings, knowledge-augmented content, and training data to augment downstream user modeling systems.}
    \label{fig:enter-label}
\end{figure}
\label{sec:approach_enhancer}

\subsubsection{Profiler}
Using LLMs-as-profilers involves the creation of prompts based on users' history, including their watching, purchasing, and viewing activities. These prompts are then inputted into LLMs to generate various aspects of users' profiles, such as their characteristics, personality, geographical location, and areas of interest. The resulting user profiles are represented in natural language, making them easily understandable to humans. They are commonly employed as input for tasks such as recommendation and rating prediction, enabling downstream predictions to be tailored to the specific requirements of individual users. HKFR \cite{yin2023heterogeneous} utilizes user heterogeneous behavior, encompassing behavior subjects, behavior content, and behavior scenarios, and feeds it into ChatGPT to obtain user profiles. ONCE \cite{liu2023once} and GENRE \cite{liu2023genre} employ LLMs to generate topics and regions of interest based on user browsing history. PALR \cite{chen2023palr} uses an LLM and user behavior as input to generate user profile keywords. KAR \cite{xi2023towards} leverages LLMs to generate user and item profiles, encompassing user preferences and factual knowledge about items, respectively. LGIR \cite{du2023enhancing} completes users' resumes by incorporating explicit properties from their self-description and implicit characteristics from their behavior history. GIRL \cite{zheng2023generative} leverages LLMs to generate suitable job descriptions based on the user's curriculum vitae to help the recommendation model better understand the job seeker's preferences. NIR \cite{wang2023zero} feeds user-watching history and design prompts for LLMs to generate user preference to augment recommendations.
Once these user profiles, which include information such as age, gender, preferences, topics of interest, and geographic location, are obtained, they can be fed into the downstream user modeling system to enhance understanding of user preferences.

\begin{table}[!t] 
\centering
\caption{Representative approaches of LLMs-as-Enhancers, LLMs-as-Controllers, LLMs-as-Evaluators.}
\label{tab:approach2}
\renewcommand\arraystretch{1.25}
\resizebox{1\linewidth}{!}{
\begin{tabular}{ccccc}
\toprule[1.5pt]
\multicolumn{1}{c|}{\makecell[c]{\textbf{Paradigms}}} & \multicolumn{1}{c|}{\makecell[c]{\textbf{Roles}}} & \multicolumn{1}{c|}{\makecell[c]{\textbf{Applications}}}& \multicolumn{1}{c|}{\makecell[c]{\textbf{LLM Backbones}}} & \multicolumn{1}{c}{\makecell[c]{\textbf{Models}}}\\ \bottomrule

\multicolumn{1}{c|}{\multirow{24}{*}{\makecell[c]{LLMs-as-Enhancers}}} & \multicolumn{1}{c|}{\multirow{4}{*}{\makecell[c]{Profiler}}} & \multicolumn{1}{c|}{\multirow{4}{*}{\makecell[c]{Recommendation}}} &  \multicolumn{1}{c|}{ChatGPT} &
\begin{tabular}[c]{@{}c@{}}
GENRE \cite{liu2023genre}, HKPF \cite{yin2023heterogeneous}, KAR \cite{xi2023towards}
\end{tabular} \\ \cline{4-5}

\multicolumn{1}{c|}{} & \multicolumn{1}{c|}{} & \multicolumn{1}{c|}{} & \multicolumn{1}{c|}{GPT-3} &
\begin{tabular}[c]{@{}c@{}}
NIR \cite{wang2023zero}
\end{tabular} \\ \cline{4-5}

\multicolumn{1}{c|}{} & \multicolumn{1}{c|}{} & \multicolumn{1}{c|}{} & \multicolumn{1}{c|}{Llama} &
\begin{tabular}[c]{@{}c@{}}
ONCE \cite{liu2023once}, PALR \cite{chen2023palr}, \citet{richardson2023integrating}
\end{tabular} \\ \cline{4-5}
\multicolumn{1}{c|}{} & \multicolumn{1}{c|}{} & \multicolumn{1}{c|}{} & \multicolumn{1}{c|}{ChatGLM} &
\begin{tabular}[c]{@{}c@{}}
LGIR \cite{du2023enhancing}
\end{tabular} \\ \cline{2-5}

\multicolumn{1}{c|}{} & \multicolumn{1}{c|}{\multirow{5}{*}{\makecell[c]{Feature Encoder}}} & \multicolumn{1}{c|}{\multirow{4}{*}{\makecell[c]{Recommendation}}} & \multicolumn{1}{c|}{GPT family} &
\begin{tabular}[c]{@{}c@{}}
GPT4SM \cite{peng2023gpt}, \citet{li2023exploring}
\end{tabular} \\ \cline{4-5}

\multicolumn{1}{c|}{} & \multicolumn{1}{c|}{} & \multicolumn{1}{c|}{} & \multicolumn{1}{c|}{Llama family} &
\begin{tabular}[c]{@{}c@{}}
LKPNR \cite{runfeng2023lkpnr}, LLM4Jobs \cite{li2023llm4jobs}
\end{tabular} \\ \cline{4-5}

\multicolumn{1}{c|}{} & \multicolumn{1}{c|}{} & \multicolumn{1}{c|}{} & \multicolumn{1}{c|}{ChatGLM} &
\begin{tabular}[c]{@{}c@{}}
LKPNR \cite{runfeng2023lkpnr}, KAR \cite{xi2023towards}
\end{tabular} \\ \cline{4-5}

\multicolumn{1}{c|}{} & \multicolumn{1}{c|}{} & \multicolumn{1}{c|}{} & \multicolumn{1}{c|}{RWKV} &
\begin{tabular}[c]{@{}c@{}}
LKPNR \cite{runfeng2023lkpnr}
\end{tabular} \\ \cline{3-5}

\multicolumn{1}{c|}{} & \multicolumn{1}{c|}{} & \multicolumn{1}{c|}{User Profiling} & \multicolumn{1}{c|}{GPT family} &
\begin{tabular}[c]{@{}c@{}}
SentimentGPT \cite{kheiri2023sentimentgpt}
\end{tabular} \\ \cline{2-5}

\multicolumn{1}{c|}{} & \multicolumn{1}{c|}{\multirow{7}{*}{\makecell[c]{\\ Knowledge Augmenter}}} & \multicolumn{1}{c|}{\multirow{4}{*}{\makecell[c]{Recommendation}}} & \multicolumn{1}{c|}{ChatGPT} &
\begin{tabular}[c]{@{}c@{}}
HKPR \cite{yin2023heterogeneous}, KAR \cite{xi2023towards}, \citet{fang2023chatgpt}
\end{tabular} \\ \cline{4-5}

\multicolumn{1}{c|}{} & \multicolumn{1}{c|}{} & \multicolumn{1}{c|}{} & \multicolumn{1}{c|}{GPT-3} &
\begin{tabular}[c]{@{}c@{}}
MINT \cite{mysore2023large}, LLM-Rec \cite{lyu2023llm}
\end{tabular} \\ \cline{4-5}

\multicolumn{1}{c|}{} & \multicolumn{1}{c|}{} & \multicolumn{1}{c|}{}  & \multicolumn{1}{c|}{GPT-2} &
\begin{tabular}[c]{@{}c@{}}
GPT4Rec \cite{li2023gpt4rec}, \citet{li2023prompt}
\end{tabular} \\ \cline{4-5}

\multicolumn{1}{c|}{} & \multicolumn{1}{c|}{} & \multicolumn{1}{c|}{}  & \multicolumn{1}{c|}{Alpaca} &
\begin{tabular}[c]{@{}c@{}}
\citet{acharya2023llm}
\end{tabular} \\ \cline{3-5}

\multicolumn{1}{c|}{} & \multicolumn{1}{c|}{} & \multicolumn{1}{c|}{Dialogue System}  & \multicolumn{1}{c|}{GPT-3} &
\begin{tabular}[c]{@{}c@{}}
TacoBot \cite{mo2023roll}
\end{tabular} \\ \cline{3-5}

\multicolumn{1}{c|}{} & \multicolumn{1}{c|}{} & \multicolumn{1}{c|}{Healthcare}  & \multicolumn{1}{c|}{GPT family} &
\begin{tabular}[c]{@{}c@{}}
PULSAR \cite{li2023pulsar}, AugESC \cite{zheng2022augesc}, \\
\citet{schlegel2023pulsar}, LLM-PTM \cite{yuan2023llm}
\end{tabular} \\ \cline{3-5}

\multicolumn{1}{c|}{} & \multicolumn{1}{c|}{} & \multicolumn{1}{c|}{Discrimination Detection}  & \multicolumn{1}{c|}{GPT-3} &
\begin{tabular}[c]{@{}c@{}}
\citet{cohen2023enhancing}
\end{tabular} \\ \cline{2-5}

\multicolumn{1}{c|}{} & \multicolumn{1}{c|}{\multirow{9}{*}{\makecell[c]{Data Generator}}} & \multicolumn{1}{c|}{Intelligent Assistant} & \multicolumn{1}{c|}{OPT-175B} &
\begin{tabular}[c]{@{}c@{}} 
VA-Model \cite{bang2022enabling}
\end{tabular} \\ \cline{3-5}

\multicolumn{1}{c|}{} & \multicolumn{1}{c|}{} & \multicolumn{1}{c|}{Recommendation}  & \multicolumn{1}{c|}{GPT family} & 
\begin{tabular}[c]{@{}c@{}}
LLM4Jobs \cite{li2023llm4jobs}, Graph-Toolformer \cite{zhang2023graph}
\end{tabular} \\ \cline{3-5}

\multicolumn{1}{c|}{} & \multicolumn{1}{c|}{} & \multicolumn{1}{c|}{Healthcare}  & \multicolumn{1}{c|}{ChatGPT} &
\begin{tabular}[c]{@{}c@{}}
\citet{tang2023does}, \citet{chen2023llm}, \citet{wang2023umass_bionlp}
\end{tabular} \\ \cline{3-5}

\multicolumn{1}{c|}{} & \multicolumn{1}{c|}{} & \multicolumn{1}{c|}{Misinformation Detection}  & \multicolumn{1}{c|}{ChatGPT} &
\begin{tabular}[c]{@{}c@{}}
\citet{su2023fake}, \citet{leite2023detecting},\citet{pan2023risk}
\end{tabular} \\ \cline{3 -5}

\multicolumn{1}{c|}{} & \multicolumn{1}{c|}{} & \multicolumn{1}{c|}{Discrimination Detection}  & \multicolumn{1}{c|}{ChatGPT} &
\begin{tabular}[c]{@{}c@{}}
\citet{veselovsky2023generating}
\end{tabular} \\ \cline{3 -5}

\multicolumn{1}{c|}{} & \multicolumn{1}{c|}{} & \multicolumn{1}{c|}{User Profiling}  & \multicolumn{1}{c|}{PaLM} &
\begin{tabular}[c]{@{}c@{}}
\citet{deng2023llms}
\end{tabular} \\ \cline{3 -5}

\multicolumn{1}{c|}{} & \multicolumn{1}{c|}{} & \multicolumn{1}{c|}{Dialogue System}  & \multicolumn{1}{c|}{GPT-4} &
\begin{tabular}[c]{@{}c@{}}
\citet{foosherian2023enhancing}
\end{tabular} \\ \cline{3 -5}

\multicolumn{1}{c|}{} & \multicolumn{1}{c|}{} & \multicolumn{1}{c|}{Fraud Detection}  & \multicolumn{1}{c|}{GPT-4} &
\begin{tabular}[c]{@{}c@{}}
\citet{yang2023anatomy}, \citet{ayoobi2023looming}
\end{tabular} \\ \cline{3 -5}

\multicolumn{1}{c|}{} & \multicolumn{1}{c|}{} & \multicolumn{1}{c|}{LLM-Gen Text Detection} & \multicolumn{1}{c|}{GPT family} &
\begin{tabular}[c]{@{}c@{}}
\citet{yu2023gpt}, \citet{chen2023can}, \citet{liu2022coco}
\end{tabular} \\ \cline{1-5}

\multicolumn{1}{c|}{\multirow{4}{*}{\makecell[c]{LLMs-as-Controllers}}} & \multicolumn{1}{c|}{\multirow{4}{*}{\makecell[c]{Pipeline Controller}}} & \multicolumn{1}{c|}{\multirow{3}{*}{\makecell[c]{Recommendation}}} & \multicolumn{1}{c|}{LaMDA} & \begin{tabular}[c]{@{}c@{}}
RecLLM \cite{friedman2023leveraging}
\end{tabular}\\ \cline{4-5}

\multicolumn{1}{c|}{} & \multicolumn{1}{c|}{} & \multicolumn{1}{c|}{}  & \multicolumn{1}{c|}{ChatGPT} &
\begin{tabular}[c]{@{}c@{}}
Chat-REC \cite{gao2023chat}
\end{tabular} \\ \cline{4-5}

\multicolumn{1}{c|}{} & \multicolumn{1}{c|}{} & \multicolumn{1}{c|}{}  & \multicolumn{1}{c|}{Vicuna} &
\begin{tabular}[c]{@{}c@{}}
LLM4Jobs \cite{li2023llm4jobs}
\end{tabular} \\ \cline{3-5}

\multicolumn{1}{c|}{} & \multicolumn{1}{c|}{} & \multicolumn{1}{c|}{Dialogue System} & \multicolumn{1}{c|}{GPT-4} &
\begin{tabular}[c]{@{}c@{}}
\citet{foosherian2023enhancing}
\end{tabular} \\ \cline{1-5}

\multicolumn{1}{c|}{\multirow{5}{*}{\makecell[c]{LLMs-as-Evaluators}}} & \multicolumn{1}{c|}{\multirow{5}{*}{\makecell[c]{Evaluator}}} & \multicolumn{1}{c|}{\multirow{3}{*}{\makecell[c]{Dialogue System}}}  & \multicolumn{1}{c|}{GPT family} &
\begin{tabular}[c]{@{}c@{}}
\citet{svikhnushina2023approximating}, \citet{huynh2023understanding},\\
\citet{zheng2023judging}, LLM-Eval \cite{lin2023llm}
\end{tabular} \\ \cline{4-5}

\multicolumn{1}{c|}{} & \multicolumn{1}{c|}{} & \multicolumn{1}{c|}{} & \multicolumn{1}{c|}{Claude} &
\begin{tabular}[c]{@{}c@{}}
\citet{huynh2023understanding}, \citet{zheng2023judging}
\end{tabular} \\ \cline{4-5}

\multicolumn{1}{c|}{} & \multicolumn{1}{c|}{} & \multicolumn{1}{c|}{}   & \multicolumn{1}{c|}{TNLG, BLOOM, OPT, Flan-T5} & 
\begin{tabular}[c]{@{}c@{}}
\citet{huynh2023understanding}
\end{tabular} \\ \cline{3-5}

\multicolumn{1}{c|}{} & \multicolumn{1}{c|}{} & \multicolumn{1}{c|}{Recommendation}  & \multicolumn{1}{c|}{\multirow{1}{*}{\makecell[c]{GPT family}}} &
\begin{tabular}[c]{@{}c@{}}
GIRL \cite{zheng2023generative}, \citet{wang2023rethinking}
\end{tabular} \\ \cline{3-5}

\multicolumn{1}{c|}{} & \multicolumn{1}{c|}{} & \multicolumn{1}{c|}{Education} & \multicolumn{1}{c|}{GPT-3} &
\begin{tabular}[c]{@{}c@{}}
\citet{bhat2022towards}
\end{tabular} \\

\bottomrule[1.5pt]

\end{tabular}
}

\end{table}

\subsubsection{Feature Encoder}
Given that LLMs have incredible user-generated content understanding and modeling capabilities, some research focuses on using LLM-generated text embeddings to enhance UM systems. GPT4SM \cite{peng2023gpt} uses both PLMs and GPT to encode recommendation queries and candidate text for relevance prediction. LKPNR \cite{runfeng2023lkpnr} uses open-source LLMs such as Llama and RWKV \cite{peng2023rwkv} to get news representation for better semantic capture capability. \citet{li2023exploring} explore LLMs' potential in text-based collaborative filtering by using LLMs with parameters ranging from 125M to 175B as feature encoders. LLM4Jobs \cite{li2023llm4jobs} leverages LLMs to embed both job posts and occupation taxonomy database and calculate their embedding similarity for recommendations. SentimentGPT \cite{kheiri2023sentimentgpt} uses GPT to encode text embeddings for small ML model training on sentiment analysis tasks. KAR \cite{xi2023towards} leverages ChatGLM as a knowledge encoder to get latent embeddings for user profiles and item factual knowledge.
These LLM-based text embeddings are then fed into the downstream models to inject the LLMs' UGC understanding ability into downstream task-agnostic models.
LLMs are shown to have strong natural language understanding ability and rich open-world knowledge. Armed with LLM-encoded embeddings, the downstream UM systems could better understand the semantic information in UGC and leverage the world knowledge from LLM latent space.

\subsubsection{Knowledge Augmenter}
LLMs have demonstrated impressive capabilities in internalizing knowledge and responding to common queries \cite{ouyang2022training,openai2023gpt4}, which can serve as knowledge bases and bring factual knowledge to UM systems. \citet{yin2023heterogeneous} leverages LLMs to fuse diverse heterogeneous knowledge, including multiple behavior subjects, multiple behavior contents, and multiple behavior scenarios. \citet{mysore2023large} augment narrative-driven recommendation using LLMs for author narrative query generation based on user-item interactions and train retrieval models with these LLM-augmented queries. KAR \cite{xi2023towards} prompts LLMs to generate factual knowledge relevant to the item for recommender system augmentation. GPT4Rec \cite{li2023gpt4rec} leverages GPT to expand users' search queries to give item titles and users' history and feed item titles for rephrasing. \citet{li2023prompt} use prompt tuning in LLMs to generate aspect embedding extraction and then feed the embeddings into aspect-based recommendation systems. LLM-Rec \cite{lyu2023llm} proposes four prompting strategies to encourage LLMs to generate knowledge-augmented item descriptions for recommendation. \citet{acharya2023llm} leverage LLMs to generate detailed item descriptions for knowledge-augmented recommendations. \citet{fang2023chatgpt} leverage ChatGPT knowledge to generate rephrases of training datasets instances to enhance model generalization and performance on unseen compositions. PULSAR \cite{li2023pulsar} integrates LLMs in data augmentation to generate more knowledge relevant to the annotated training data. Knowledge stored in LLMs can also be used to augment dialogue. AugESC \cite{zheng2022augesc} uses LLMs to complete full dialogue and construct a scalable dataset to augment dialog systems' generalization ability in open-domain topics. \citet{schlegel2023pulsar} ask LLMs to generate a hypothetical conversation between the doctor and the patient based on a medical note and then use this data to train a specialized LM. \citet{cohen2023enhancing} leverage GPT-3 to do back translation and rephrase as data augmentations for hate speech detection. TacoBot \cite{mo2023roll} leverages LLMs to synthesize user intents of conversation for data augmentation. LLM-PTM \cite{yuan2023llm} proposes an LLM-based privacy-aware augmentation for patient-trial matching, which leverages LLMs to create supplementary data points while preserving the semantic coherence of the original trail's inclusion and exclusion criteria. 
The collective findings of these works show that researchers could enhance UM systems performance by identifying and injecting external reasoning or factual knowledge into UM text input. Take movie recommendation as an example, LLMs provide contextual knowledge for candidate item descriptions under specific scenarios. This approach paves the way for the advancement of open-world recommendation, which integrates broader contextual information to enhance the recommender system.

\subsubsection{Data Generator}
Owe to LLMs' strong sequence generation and data compression ability \cite{yu2023large}, there is a lot of research that leverages LLMs to generate training data or weak labels and feed these into small UM models for training \cite{peng2023generating, borisov2022language, viswanathan2023prompt2model}, which can also be envisioned as knowledge distillation. 
VA-Models \cite{bang2022enabling} generates value-aligned training data from LLMs by few-shot prompting, then uses the generated data to train small models on value-alignment judgment tasks. 
\citet{tang2023does} investigates LLMs' potential to synthesize high-quality clinical data with labels and fine-tune local models for the downstream tasks. 
LLM4Job \cite{li2023llm4jobs} leverages GPT-4 to generate occupation coding datasets. \citet{su2023fake} investigate fake news detection models' bias with LLMs-synthesized fake news articles. \citet{veselovsky2023generating} find LLMs-synthesized data have a different distribution than real-world data, and propose three strategies: ground, filter, and taxonomy-based generation strategy to combat this difference, which has proved to be effective in sarcasm detection. LLMs can also generate weak labels to enhance UM systems. For example, \citet{leite2023detecting} propose to prompt LLMs with credibility signals to produce weak labels to enhance misinformation detection performance. \citet{deng2023llms} propose to generate weak financial sentiment labels for Reddit posts with an LLM and then use that data to train a small model that can be served in production. 
\citet{foosherian2023enhancing} find that LLMs can aid conversational agents in generating training data, extracting entities and synonyms, localization, and persona design. For the LLM-generated text detection domain, LLMs are employed to synthesize training corpus \cite{yu2023gpt, chen2023can, liu2022coco, yu2023pre}. Graph-Toolformer \cite{zhang2023graph} uses ChatGPT to annotate and augment a large prompt dataset that contains API calls of external reasoning tools and uses the synthesized dataset to fine-tune open-source LLMs. RefGPT \cite{yang2023refgpt} uses LLMs to generate truthful and customized dialogues without hallucination.

LLMs can also generate high-quality conversational data. \citet{zheng2023building} recursively prompting ChatGPT with in-context learning to generate an extensive emotional support dialogue dataset (ExTES) and use it to fine-tune Llama for optimized emotional support dialogue systems. \citet{wang2023umass_bionlp} propose a LLMs cooperation system named a doctor-patient loop to generate high-quality conversation datasets. 
Overall, these works indicate that by training on LLM synthesized data, small UM models could inherit the strong user understanding ability from LLMs. These approaches are advantageous in low-resource circumstances or some applications that need efficient deployment. For instance, in healthcare applications where privacy concerns and data scarcity are prevalent, utilizing LLMs to generate supplementary data can significantly alleviate the low-resource and privacy problems and thus bolster the effectiveness of UM systems.

\begin{figure}[t]
    \centering
    \includegraphics[width=1\linewidth]{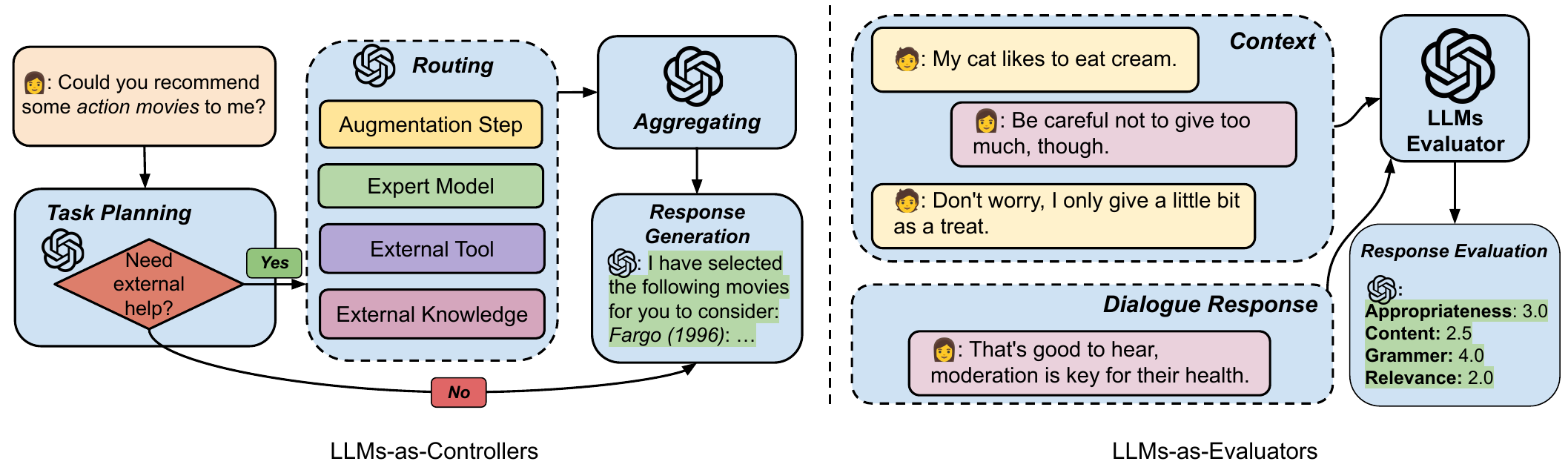}
    \vspace{-0.3in}
    \caption{LLMs-as-Controllers and LLMs-as-Evaluators, where LLMs are employed to manage the LLM-UM pipeline and to evaluate the output of UM systems, respectively.}
    \label{fig:enter-label}
\end{figure}

\subsection{LLMs-as-Controllers}
\label{sec:approach_controller}
The scale of parameters of LLMs brings emergent abilities that have never been observed in small language models and gives LLMs unprecedented ability to control the system pipeline and supercharge the UM system for personal needs. Note that different from LLMs as agents that let LLMs freely explore and interact with the environment, LLMs-as-Controllers include works that have designed the entire pipeline and let LLMs decide whether to conduct certain operations.
HuggingGPT \cite{shen2023hugginggpt} employs LLMs as a controller to manage and organize the cooperation of expert models. Specifically in the user modeling systems, RecLLM \cite{friedman2023leveraging} leverages LLMs as a dialogue manager, which converses with the user, tracks context, and makes system calls when necessary. Chat-REC \cite{gao2023chat} lets LLMs decide when to use recommendation systems as external tools. \citet{foosherian2023enhancing} demonstrate that LLMs can assist the pipeline-based conversational agent in contextualization, intent classification to prevent conversational breakdown and handle out-of-scope questions, auto-correcting utterances, rephrasing responses, formulating disambiguation questions, summarization, and enabling closed question-answering capabilities. LLM4Jobs \cite{li2023llm4jobs} constructs a pipeline that makes LLMs determine whether to do a summary on the job posting.

\subsection{LLMs-as-Evaluators}
Evaluating natural language generation (NLG), especially in the open-domain and conversation settings, has posed significant challenges in user modeling. The strong language modeling capabilities in LLMs open up new opportunities for these complex evaluations, and some research works propose to envision LLMs as evaluators for generative UM systems.
\citet{svikhnushina2023approximating} leverage LLMs to approximate online human evaluation for dialogue systems. \citet{huynh2023understanding} explores LLMs capabilities in dialog evaluation and the relation between prompts and training datasets. iEvaLM \cite{wang2023rethinking} proposed a conversational recommendation evaluation framework that leverages LLMs to simulate various interactions between users and systems. \citet{zheng2023judging} discover that using strong LLMs like GPT-4 as a judge can match both controlled and crowdsourced human preferences well. LLM-Eval \cite{lin2023llm} designs a prompt-based evaluation method that leverages a unified evaluation schema to cover multiple dimensions of conversation quality. \citet{bhat2022towards} takes a fine-tuned GPT-3 to evaluate the generated questions by classifying if the questions are useful to learning. GIRL \cite{zheng2023generative} evaluates the recommended job results with the help of ChatGPT.
These works reveal that LLMs can be an effective tool of assessing UM system output, determining the extent these outputs are customized to meet specific user needs. Particularly in conversational contexts, where conducting user studies could be expensive and prone to bias, LLMs provide a reliable and effective approach to assess the quality of complex and open-ended generations. Consequently, the LLMs-as-Evaluator paradigm enhances UM system development.

\begin{mybox}
\label{personalizeprompting}

\paragraph{\textit{Personalized Prompting}}

It is worth mentioning that to make LLM-UM adapt to an individual's specific needs and generate personalized content, existing research works design prompting templates to encode UGC and user behavior history to help LLMs understand users' preferences. Existing personalized prompting paradigms fall into three categories: vanilla, retrieval-augmented, and profile-augmented methods. We provide specific examples of these prompts in the context of the personalized product rating task below.

\begin{itemize}
\item \texttt{\textbf{Vanilla Personalized Prompt:}} Here is the user rating history: \{\{\texttt{all user behavior history}\}\}. Based on the above rating history, please predict the user's rating for the product: \{\{\texttt{query item}\}\}.
\item \texttt{\textbf{Retrieval-Augmented Personalized Prompt:}} Here is the user rating history: \{\{\texttt{top-k user behavior history}\}\}. Based on the above rating history, please predict the user's rating for the product: \{\{\texttt{query item}\}\}.
\item \texttt{\textbf{Profile-Augmented Personalized Prompt:}} \{\{\texttt{User Profile}\}\}. Here is the user rating history: \{\{\texttt{top-k user behavior history}\}\}. Based on the above rating history, please predict the user's rating for the product: \{\{\texttt{query item}\}\}.
\end{itemize}

Most works under the LLMs-as-Predictors paradigm adopt the in-context learning paradigm and encode the entire user behavior history as in-context examples (\textbf{Vanilla Personalized Prompt}). BookGPT \cite{zhiyuli2023bookgpt}, for instance, employs a few-shot prompting strategy to enable LLMs to understand the correlation between book content and personalized ratings. \citet{dai2023uncovering} encode user rating history as few-shot demonstration examples. \citet{liu2023chatgpt} supply the LLM with the user's interaction history relating to the task, enabling the generation of personalized content. \citet{christakopoulou2023large} utilize few-shot prompting to demonstrate the user research journey to LLMs.
and find that feeding partial user behavior log causes significantly lower performance. Moreover, BookGPT \cite{zhiyuli2023bookgpt} discovers that a longer user behavior history would bring better performance.
Considering the increasing length of UGC and user behavior history and the limited context length of LLM, some studies have applied retrieval methods to select the most relevant user data for enhancing LLM personalization (\textbf{Retrieval-Augmented Personalized Prompt}). For instance, LaMP \cite{salemi2023lamp} introduces a retrieval-augmented method to obtain the most relevant content in behavioral history and incorporate it into the prompt.

Except for previously mentioned works that fall into the LLMs-as-Predictors paradigm, some works can be categorized into the LLMs-as-Enhancers paradigm that employs LLMs to rephrase and summarize user preferences and profiles based on their history (\textbf{Profile-Augmented Personalized Prompt}). \citet{richardson2023integrating} suggest the use of instruction-tuned LLMs to generate an abstract summary of user data, augmenting retrieval-based personalized methods like LaMP. HKFP \cite{yin2023heterogeneous} inputs user behavior history into LLMs to fuse heterogeneous user knowledge, which assists fine-tuned LLMs in understanding user preferences. Note that the generated profile could vary in different tasks. For example, in rating prediction tasks, \citet{richardson2023integrating} prompt instruction-tuned LLMs to summarize the user's most common positive and negative score as profile, while generating user research interest as profile in the personalized citation identification task.

\end{mybox}

\section{Applications of LLM-UM}
\label{sec:application}
We introduce the applications that can be categorized into personalization and suspiciousness detection.

\subsection{Personalization}
\label{sec:app_personalization}
Personalization in user modeling refers to tailoring and customizing a system's interactions, content, or recommendations to meet the specific needs, preferences, and characteristics of individual users. In this section, we look into LLM-UM for personalization applications, including user-generated content (UGC) analysis, user behavior prediction, personalized assistance, personalized recommendation, personalized dialog system, personalized education, and personalized healthcare.


\subsubsection{User Profiling}
User profiling refers to mining the characteristics, personality, and potential preferences based on UGC and user behavior, paving the way for downstream personalized applications. 
User profiling mainly includes detecting users' stances and sentiments and analyzing users' characteristics, personalities, etc. \citet{zhang2022would} discuss LLMs' impact in stance detection topics and the opportunities they bring. LoT \cite{hu2023ladder} is proposed to help LLMs assimilate high-quality external knowledge to boost stance detection. \citet{mu2023navigating} and \citet{ziems2023can} comprehensively investigate LLMs in the computational social science tasks, including detecting sarcasm, hate speech, ideology, stance, and more in UGC. SentimentGPT \cite{kheiri2023sentimentgpt} is a pioneering to leverage LLM for UGC sentiment analysis. \citet{wu2023large} discover that LLMs can be used to estimate the latent stance of politicians and give solutions to complex social science measurement problems.
For general personality modeling, \citet{rao2023can} use LLMs to assess human personality based on the MBTI test. \citet{ji2023chatgpt} employ a level-oriented prompting strategy to analyze the user's personality in a given text. 
\citet{christakopoulou2023large} use LLMs to mine users' interest journey and provide deeper, more interpretable, and controllable user understandings.



\subsubsection{Personalized Recommendation}
\label{sec:per_recsys}
LLMs are adopted in the recommendation system to offer personalized suggestions towards candidate items tailored to meet user preferences and specific needs. The recommendation tasks can be further categorized into top-k recommendation, rating prediction, and conversational recommendation.

\paragraph{Top-k Recommendation}
The top-k recommendation task directly predicts the top-k favorite items based on the user's behavior history. Most methods directly design prompts feeding user behavior history and optional characteristics into LLMs to generate recommended items. In context learning is the most common paradigm for the top-k recommendation, which gives several exemplars and recommendation results to help LLMs better understand tasks. Representative works include \cite{liu2023chatgpt, dai2023uncovering, zhang2023chatgpt, liu2023genre, hou2023large, wang2023rethinking, du2023enhancing}. \citet{zhang2023recommendation} further utilized Chain-of-Thought prompting to conduct a top-k recommendation task. For better representation and domain adaptation, researchers also fine-tune LLM's parameters. \citet{chen2023palr} and GenRec \cite{ji2023genrec} fine-tune a Llama-7B model to help the model adapt to recommender system and generate items. \citet{zhang2023recommendation} conduct instruction tuning on a Flan-T5-XL model to help it adapt to recommendation. GIRL \cite{zheng2023generative} further proposes a reward model and uses reinforcement learning to provide better feedback for LLMs fine-tuning. 

\paragraph{Rating Prediction}
The rating prediction means predicting the user's rating for given items. In this process, LLMs give the predicted scores for items in the context of the user's behavior and UGC history. The rating prediction task can probe LLM's capabilities in understanding user preference and can also be understood as user behavior prediction. Similar to top-k recommendation, rating prediction can also be categorized into frozen LLMs and fine-tuning LLMs. Armed with frozen LLMs, BookGPT \cite{zhiyuli2023bookgpt} and  \citet{dai2023uncovering} conduct prompt engineering to make LLMs generate predicted ratings; KAR \cite{xi2023towards} takes LLMs to generate user profile and factual knowledge of items and feed them into the discriminative recommendation for rating prediction. For rating prediction with fine-tuned LLMs, \citet{kang2023llms} fine-tune LLMs in the rating prediction task; TallRec \cite{bao2023tallrec} integrate LoRA and instruction tuning for rating prediction task adaptation; GLRec \cite{wu2023exploring} incorporates meta-paths into soft prompt and conduct instruction tuning for item generation. Further, Graph-Toolformer \cite{zhang2023graph} fine-tunes LLMs and empowers them to use external graph reasoning tools for user rating prediction.

\paragraph{Conversational Recommendation}
Conversational recommendation means using the system to interact with the user and understand the user's preference through conversation and then generate recommended items in the conversation. Chat-REC \cite{gao2023chat} converts the user profile and historical interactions into prompts to build conversational recommendation systems. GeneRec \cite{wang2023generative} adopts ChatGPT to
personalize content generation, and it leverages user instructions to acquire users’ information needs. \citet{lin2023sparks} envision ChatGPT as an Artificial General Recommender (AGR) that comprises conversationally and universality to engage in natural dialogues and generate recommendations across various domains.

\subsubsection{Personalized Assistance}
Personalized assistance refers to using LLM techniques to tailor and customize generated content based on individual preferences, behavior, and characteristics of users. \citet{tian2023chatgpt} empirically study the LLMs' potential as fully automated programming assistants in the tasks of code generation, program repair, and code summarization. TacoBot \cite{mo2023roll} is an LLM-augmented task-oriented dialogue system that guides users through complex real-world tasks with multiple steps. LaMP \cite{salemi2023lamp} uses LLMs to conduct personalized text classification and personalized text generation, such as personalized citation identification and personalized news headline generation, etc. DISC-LawLLM \cite{yue2023disc} is an intelligent system that utilizes LLMs to provide a wide range of personalized legal advice. FinGPT \cite{liu2023fingpt} is an LLM that can offer personalized investment advice based on the user's risk and financial goals. \citet{chakrabarty2023creativity} investigate LLMs' ability in creative writing assistance, including planning, translating, and reviewing processes.

\subsubsection{Personalized Dialogue System}
LLMs are also widely adopted in dialogue systems and combined with the user's behavior history and preference to provide personalized user experiences.  
\citet{hudevcek2023llms} utilized LLMs to retrieve the user behavior context and user history for personalized conversational response generation.
DiagGPT \cite{cao2023diaggpt} extends LLMs to task-oriented dialogue scenarios, in which LLMs need to pose questions and guide users toward specific task completion proactively.
\citet{cho2022personalized} use GPT-2 to generate dialogue data while detecting the user's persona. 
 RefGPT \cite{yang2023refgpt} proposes to generate enormous truthful and customized dialogues without worrying about factual errors caused by the model hallucination to enable the personalized dialogue system.

\subsubsection{Personalized Education}
\label{sec:per_edu}
LLMs have shown great potential in promoting the equality of education and improving the existing education paradigm by adapting LLM-based tools to tailor for students and instructors \cite{joshi2023let, yan2023practical}, and education for data scientists \cite{tu2023should}. \citet{koyuturk2023developing} discover that with multiple interaction turns, LLMs can adapt the educational activity to the user's characteristics, such as culture, age, and level of education, and its ability to use diverse educational strategies and conversational styles. EduChat \cite{dan2023educhat} is an LLM-based chatbot that aims to strengthen personalized, fair, and compassionate intelligent education, serving teachers, students, and parents. \citet{sharma2023performance} investigate ChatGPT's performance on the United States Medical Licensing Examination (USMLE) and point out that ChatGPT can be an invaluable tool for e-learners. \citet{elkins2023useful} investigate ChatGPT's performance in generating educational questions in classroom settings and find them high-quality and sufficiently useful. \citet{ochieng2023large} delve into LLM's ability to participate in educational guided reading, including generating questions based on the input text and recommending content based on the user's response. C-LLM \cite{olga2023generative} examines the implications for LLMs for AI review and assessment of complex student work. 
\citet{phung2023generative} comprehensively investigate ChatGPT's capability in a set of programming education scenarios and find GPT-4 comes close to human tutor in several scenarios.

\subsubsection{Personalized Healthcare}
\label{sec:per_health}
LLMs also play an important role in empowering healthcare services and providing personalized service. \citet{liu2023large} discover that LLMs are capable of grounding various physiological and behavioral time-series data and making meaningful inferences under the few-shot settings. \citet{wang2023large} investigates the performance of LLMs on clinical language understanding tasks and introduces self-questioning prompting to enhance LLM in clinical scenarios. 
HeLM \cite{belyaeva2023multimodal} enables LLMs to use high-dimensional clinical modalities to estimate underlying disease risk with individual-specific data. 
PharmacyGPT \cite{liu2023pharmacygpt} utilizes LLms to generate comprehensible patient clusters, formulate medication plans, and forecast patient outcomes. \citet{zhang2023potential} utilize LLMs to identify patients with specific medical diagnoses and provide diagnostic assistance to healthcare workers. Zhongjing \cite{yang2023zhongjing} introduces a Llama-based LLM with expertise in the Chinese medical domain and can provide personalized advice for the user's specific case.

LLMs are widely applied for mental health. \citet{ghanadian2023chatgpt} utilize LLMs for suicidality assessment from social media. \citet{qi2023evaluating} takes LLMs to evaluate suicidal risk and cognitive distortion identification on Chinese social media platforms. \citet{wang2023umass_bionlp} leverage LLMs to generate note-oriented doctor-patient conversations. \citet{chen2023llm} utilize ChatGPT to simulate conversations between psychiatrist and patient based on user experience. \citet{fu2023enhancing} present an LLM-empowered framework that assists non-professionals in providing psychological interventions on online user discourse. Mental-LLM \cite{xu2023mental} investigate LLM's capabilities in mental health tasks and find the superiority of instruction tuning in boosting LLMs' performance in mental health tasks. \citet{lai2023psy} propose an LLM-based assistant for question-answering in psychological consultation settings to ease the demand for mental health professions. \citet{peters2023large} assess the ability of GPT-3.5 and GPT-4 to infer the psychological dispositions of individuals from their digital footprints.


\subsection{Suspiciousness Detection}
\label{sec:app_suspicious}
User modeling involves the process of understanding and predicting users' behavior and preferences. By creating a comprehensive model of user's normal behaviors, it becomes possible to detect deviations from this norm. For users with suspicious behavior history, UM system could potentially isolate or warn users who exhibit harmful behaviors, while offering a more open environment to those with positive engagement histories. Therefore, suspiciousness detection is a key application of user modeling.
Suspiciousness detection refers to the process of identifying or recognizing behaviors, actions, or patterns that are deemed to be suspicious or potentially indicative of anomalous, illegal, harmful, or malicious activities \cite{zhao2021action,jiang2014detecting,jiang2014catchsync}. 
This section introduces leveraging LLM-UM to detect fraud, hate speech, misinformation, misconduct, and LLM-generated content. 

\subsubsection{Fraud Detection}
\label{sec:sus_fraud}
Suspiciousness in fraud detection refers to fraudulent and deceptive behavior, which is widely adopted in financial transactions, social networks, and more. Some research leverages LLMs to supercharge fraud detection models. \citet{zhu2023clickbait} design prompts to enable LLMs for clickbait detection and achieves satisfied performance. \citet{yang2023anatomy} present a case study on a Twitter botnet that employs ChatGPT to generate human-like content, while state-of-the-art LLM content classifiers fail to detect them. Spam-T5 \cite{labonne2023spam} fine-tunes T5 for spam email detection and outperforms baselines with limited training samples. \citet{ayoobi2023looming} leverage LLMs to generate fake LinkedIn profiles and develop the Section Tag Embeddings to detect fake profiles. \citet{shukla2023catch} explores GPT-3 and GPT-4 for fraudulent physician review detection.
\citet{ziems2023explaining} employs GPT-4 to explain the decisions of classicial machine learning models on network intrusion detection.

\subsubsection{Discrimination Detection}
\label{sec:hatespeech}
LLMs' strong natural language understanding ability can supercharge hate speech detection, which is of great importance in social content moderation \cite{matwin2021survey}. \citet{chiu2021detecting} leverage GPT-3 to identify sexist and racist UGC under zero-shot, one-shot, and few-shot settings. \citet{cohen2023enhancing} leverage GPT-3 to back translate and rephrase as augmentations for hate speech detection. 
\citet{del2023respectful} explores using LLMs with prompting for zero-shot hate speech detection. \citet{das2023evaluating} evaluate LLMs' capabilities in multilingual and emoji-based hate speech. \citet{wang2023evaluating} further prompt LLMs to generate an explanation for hateful and non-hateful content and investigate the explanation generated by LLMs. \citet{nguyen2023fine} fine-tune Llama-7B to detect online sexual predatory chats and abusive language.

\subsubsection{Misinformation Detection}
\label{sec:misinfo}
LLMs can also be leveraged to detect misinformation, especially fake news \cite{jiang2023disinformation}. \citet{chen2023can} discover that LLM-generated misinformation can be harder to detect compared to human-written, which suggests a more deceptive style and potentially causes more harm. \citet{pavlyshenko2023analysis} explores using a fine-tuned Llama-2 model for misinformation analysis and fake news detection. \citet{yang2023large} assess ChatGPT's ability to rate the credibility of news outlets and find ChatGPT's prediction correlates to those from human experts. \citet{pan2023risk} establish a threat model and reveal that LLMs can act as effective misinformation generators, leading to significant degradation in open-domain question-answering systems. \citet{leite2023detecting} develop a misinformation detection approach that combines the zero-shot LLM credibility signal labeling and weak supervision and achieve state-of-the-art without using ground truth label for training. \citet{su2023fake} discover that existing detectors are prone to flagging LLM-generated text as fake news and propose to leverage adversarial training for bias mitigation. \citet{huang2023harnessing} explore ChatGPT's proficiency in generating, explaining, and detecting fake news, and propose using potential extra information that could boost LLM-based fake news detection.


\subsubsection{LLM-Generated Text Detection}
\label{sec:sus_LLMgen}
As LLMs emerge, misuse of LLMs increases, including disseminating fake news, plagiarism, manipulating public opinion \cite{hanley2023machine}, cheating, and fraud, making detection of LLM-generated content essential \cite{pegoraro2023chatgpt, dhaini2023detecting, tang2023science, chen2023can, lu2023large}.
Researchers find that LLM-generated text can easily evade the plagiarism-checking software \cite{khalil2023will} and is difficult to be identified by humans \cite{wahle2022large}. SpaceInfi \cite{cai2023evade} discovers that the extra space can serve an important role for LLM-generated content to evade detection.

GPT-Pat \cite{yu2023gpt} develops a framework consisting of a Siamese network to compute the similarity between original and LLM-generated text and a binary classifier. \citet{yang2023chatgpt} measure the ChatGPT's involvement in text generation based on edit distance. \citet{orenstrakh2023detecting} investigate the LLM-generated text detection tool specifically in the education domain. CoCo \cite{liu2022coco} presents a coherence-based contrastive learning method to detect LLM-generated text under the low-resource scenario. \citet{bhattacharjee2023fighting} employ ChatGPT as a detector for AI-generated text detection. 


\section{Challenges and Directions}
\label{sec:future}
In this survey, we have comprehensively reviewed the recent approaches and applications of LLM-UM. Since the integration of LLMs into user modeling systems is still in the early stage, there are still some important and challenging yet unsolved problems in this direction. In this section, we discuss some challenges and potential future directions in this field.

\subsection{Hallucination Mitigation}
Though LLMs have shown strong capabilities in various domains, a significant challenge is the hallucination problem \cite{zhang2023siren, ji2023survey}, where LLMs generate seemingly plausible content but deviate from user input \cite{adlakha2023evaluating}, previously generated content \cite{liu2021token}, and factuality knowledge \cite{min2023factscore, feng2023factkb}. In the context of user modeling, the challenges of hallucination can be categorized into 1) \textit{factual accuracy}: LLMs can sometimes generate content that sounds plausible but is either factually incorrect or not grounded in the input data. 2) \textit{hallucination in high-stakes scenarios}: In contexts where recommendations have profound implications, such as in healthcare or legal advice, hallucinations can have severe consequences. 3) \textit{user intent misunderstanding}: Hallucination can also arise from a misunderstanding of user intent and input context, which can significantly hinder the performance of LLMs in user modeling.



To mitigate the prevalent hallucination problems in LLMs, a promising approach to counter the hallucination problem in LLM-UM is to incorporate trustworthy symbolic knowledge in the LLMs' input, including unifying knowledge graphs (KGs) \cite{pan2023unifying} and the use of external knowledge retrieval for augmented generation \cite{yu2022survey}. Manipulating the output side of LLMs can also help alleviate the hallucination problem. This could involve generating calibrated confidence scores to validate predictions and encouraging LLMs to express uncertainty. Research could also focus on designing decoding strategies that foster more reliable content generation, including factuality feedback mechanisms for the next token selection.

\subsection{Privacy and Security}
Privacy and security are paramount concerns in the user modeling system since it can involve processing and analyzing a wealth of UGC and user behavior history. However, recent research suggests that LLMs can pose privacy risks. For example, an adversary can recover training examples containing a person’s name, email address, and phone number by querying the model \cite{huang2022large}. The challenges in LLM-UM can be summarized as 1) \textit{balancing personalization and privacy}: The success of LLMs in user modeling relies heavily on personalization, which necessitates the use of sensitive user data. However, this can lead to privacy and security concerns. That being said, striking the right balance between personalization and privacy is a significant challenge. 2) \textit{data leakage risk}: As LLMs are trained on large quantities of data, there's a risk that they might unintentionally leak sensitive information during the text generation process. Recent research has shown that LLMs are vulnerable to adversarial attacks by prompt injection \cite{li2023you} and jailbreaking \cite{rao2023tricking}. This could potentially leak users' identities or private information and cause security issues. 3) \textit{adaptation to evolving privacy laws}: Privacy laws and regulations are continually evolving, and LLMs need to be updated to ensure compliance, which can be a challenging task given the huge parameters and high updating cost.

To mitigate the above challenges, future research can focus on designing privacy techniques in preprocessing, training, and post-processing steps to mitigate user privacy risks. They can study how to provide users with more transparency and study how their data are used and give them more control over their information.


\subsection{Complex Structural Data Understanding}
User interactions fundamentally form complex, heterogeneous, temporal text-rich graphs, encapsulating user-user and user-item interactions. Therefore, equipping LLMs with the capability to comprehend and interpret such graph structures could dramatically improve the performance of user modeling. Current challenges of leveraging LLMs to model complex graph structure can be summarized as 1) \textit{lack of sequential transformation}: Unlike natural languages, graph-structured data often lack a straightforward transformation into sequential text \cite{wang2020calendar}, which makes it difficult for LLMs to process and understand. 2) \textit{gap between graphs and human languages}: The significant gap between graph structures and natural language used for LLMs pretraining can hinder the application of LLMs to facilitate graph reasoning. 3) \textit{graph variability}: Graphs can define structure and features in distinct ways and lack a unified paradigm, making it hard to generalize well to the diverse structure and feature representations of different graphs \cite{jiang2016catchtartan,wang2021modeling}. 4) \textit{heterogeneous temporal text-rich graph structure modeling}: User interactions can be formulated into heterogeneous temporal text-rich graphs, posing challenges for LLMs to understand such complex data.

To empower LLMs with the capabilities to understand complex structural data, future research could focus on 1) \textit{graph language generation}: One potential research direction could be to derive a language for graph structures that LLMs can better understand, such as transforming graph data into a form that LLMs can process, such as sequential text. 2) \textit{temporal heterogeneous graph modeling}: Given that user interactions often form heterogeneous temporal graphs, the research could focus on techniques to handle such complex structures, which could involve developing methods to capture the dynamics and heterogeneity of user interactions.

\subsection{Comprehensive Benchmark and Evaluation Criteria}
Evaluation is paramount to the success of LLM-UM, which can help us better understand the strengths and weaknesses, ensuring safety and reliability, and inspiring future research directions of user modeling systems.
 Current challenges of LLM-UM evaluations lie in 1) \textit{lack of AGI-level benchmark}: Current evaluation benchmarks mainly focus on restricting models input and output to a static pair based on human annotation, which cannot measure the superhuman capabilities in LLMs, instead, the existing benchmarks can be seen in LLMs pertaining stage and cause test data leakage problem. Therefore, designing more challenging tasks and more reliable evaluation mechanisms by utilizing the cross-disciplinary knowledge from education, social science, and more could be a better solution to probe LLM-UM's capabilities as a superhuman-level system.
2) \textit{comprehensive intermediate evaluation}: A general concern in LLM-UM is the evaluation only focuses on the prediction performance on specific datasets. Although this is likely the most important metric, the intermediate steps such as user profile generation, LLM-generated factual knowledge and datasets, and pipeline control behavior are also important for researchers to more comprehensively understand the methods.
3) \textit{dynamic and evolving evaluation}: Existing UM evaluation protocols rely on static and public datasets, which would lead to overfitting and could not assess the LLM-UM in evolving circumstances, especially in the context of fast-evolving user data. For example, AgentBench \cite{liu2023agentbench} is a pioneering work that designs evolving benchmarks and evaluates LLMs as agents, which is shown to be the better way of static input-output pair evaluation to evaluate LLM-based systems. Therefore, developing a dynamic and evolving testbed for LLM-UM to probe the system's capabilities in adapting the evolving data is an important research direction.

\subsection{Trustworthy User Modeling}
LLM-UM can bring significant benefits to humans, including providing personalized advice, recommendations, assistance, and more. However, the black-box property and unreliable generation process of LLMs would pose a serious threat to users when the LLM-UM system is applied to high-stakes domains such as healthcare and finance. Therefore, how to enable trustworthy LLM-UM is of significant importance. Current challenges and potential future research directions include 1) \textit{transparent and explainable reasoning}: Users need to understand how an LLM-UM system makes decisions to enhance trust. Therefore, promising research directions are how to explain the black-box LLMs and how to make the user modeling process transparent and explainable.
2) \textit{reliability}: LLM-UM should provide consistent and accurate results, be robust to different input templates, and give consistent generation given the same or similar input to be trustworthy. 
3) \textit{user control}: Trustworthy LLM-UM systems should allow users to control how their data is used and how decisions are made, which could involve developing mechanisms for users to feedback, correct, and update the LLM-UM system behavior.

\subsection{Fairness in LLM-UM}
Fairness is a crucial aspect of user modeling systems, which secure unbiased user characteristics and profile generation and give fair assistance and recommendations for different groups of people. LLMs trained on extensive datasets can inadvertently learn and perpetuate human biases and stereotypes in training data, leading to unfair treatment or discriminatory predictions of specific user groups \cite{feng2023pretraining}. Moreover, balancing personalization and fairness is a significant problem, especially when user data reflects societal biases. Therefore, future LLM-UM research may include 1) \textit{bias detection and measurement}: Detect bias that exists in LLM-UM systems training data, input, model, and task design, and develop fine-grained mechanisms to measure the bias from different aspects (e.g., political bias, ethnic groups bias, etc.) in LLM-UM systems. 2) \textit{promote fairness in LLM-UM systems}: Mitigate bias and give non-discriminative predictions, including making LLM-UM more explainable and transparent to help better understand the source of unfair behavior and devise appropriate interventions. 3) \textit{ethics and regulations}: Research should also focus on the ethical implications of LLMs and the potential for regulations to ensure fairness, which could involve interdisciplinary work, bringing together technologists, ethicists, sociologists, and lawmakers.


\subsection{Efficient and Effective Domain Adaptation}
LLMs are pretrained on a large general-purpose corpus and then plugged into user modeling systems. A lot of works use LLMs as a general-purpose reasoner with frozen pertaining parameters, while others fine-tune LLMs to help them adapt to the specific task in user modeling. The latter paradigm allows the model to leverage the general language understanding capabilities learned during the pretraining stage while specializing the knowledge to the task at hand and shows superior task performance. However, fine-tuning and deployment can be highly computationally expensive, especially when deploying on millions and even billions of scale user data. Therefore, developing efficient domain adaptation methods to reduce the cost for LLMs to fix in user modeling systems and efficiently help LLM-UM be personalized and adapt to the user's dynamic needs. Moreover, user-specific data is often sparse and noisy \cite{koren2009matrix, das2018personalized}, which can lead to overfitting during the fine-tuning process, which can limit the model's ability to generalize to different users and contexts. Therefore, future research to combat these challenges can focus on 1) \textit{efficient fine-tuning for LLM-UM}: There is a need to develop and refine techniques for more efficient fine-tuning of LLMs specific for user modeling context. Techniques such as sparsity induction, model pruning, and knowledge distillation could be explored to reduce the computational and memory footprints. 2) \textit{online learning}: To cope with the dynamics of user interests, research could explore online learning strategies for LLM-UM systems. These strategies would continuously update the model based on the latest user-generated content and interactions.

\subsection{Personalized LLM-UM Deployment}
Recent research on personalized LLMs mainly focuses on two aspects: personalized prompt engineering and personalized preference alignment. For personalized prompt engineering works, researchers focus on designing prompts to help LLMs understand user preferences based on user behavior. As previously discussed, the personalized prompting paradigm would directly encode user behavior history, apply retrieval methods to obtain the most relevant user history, and leverage LLMs to summarize user preference to augment retrieval and generation performance. At the same time, the other line of work focuses on aligning LLMs with personal preferences using reinforcement learning. Personalized Soups \cite{jang2023personalized} is a representative work that decomposes user preference into multiple dimensions and models the personalization into a multi-object reinforcement learning problem. However, the existing personalized prompting model exhibits limited generalization and context windows for encoding and comprehending user preferences from extensive user behavioral history. Furthermore, human alignment methods require explicit human preferences and pre-established preference dimensions, and the alignment finetuning process is computationally expensive. Therefore, future personalized LLM research should focus on 1) \textit{parametric user preference mining}: As previous studies \cite{liu2023gpt} have indicated, differential parameters are more effective in understanding and modeling user preferences compared to discrete natural language. Additionally, parametric personalization can infuse personalized knowledge more efficiently by using fewer tokens, thereby conserving the context length for LLMs.
2) \textit{cross-model personalized knowledge transfer}: Personalized LLMs strive to diverge from the prevailing `one-size-fits-all' paradigm, aiming to offer unique services tailored to each individual. However, given the `scaling law' - which posits that larger models yield superior performance - not all users possess the computational resources to generate high-quality output by operating their own large models on local machines. Therefore, future research should work towards developing small models that can learn from user history and gain personalized knowledge on a local machine. These models should be able to transfer or plug their knowledge into a remote, large model to produce high-quality personalized output.
3) \textit{parametric privacy preservation}: Recent findings suggest that encoded text embeddings alone can be used to reconstruct full text \cite{morris2023text}, posing potential privacy and safety risks. This becomes particularly relevant in personalization, where user preferences can be encoded using a small model on local machines. There exists a risk of private data reconstruction using latent embedding and parameters. Therefore, developing strategies to thwart such reconstruction while preserving latent embedding performance is a pressing issue.

\section{Conclusion}
\label{sec:conclusion}
Our work presents a comprehensive and structured survey of large language models for user modeling (LLM-UM). We show why LLMs are great tools for user modeling and understanding user-generated content (UGC) and user interactions. We then review the existing LLM-UM research works and categorize their approaches that integrate text-based and graph-based UM techniques, including LLMs serving as enhancers, predictors, controllers, and evaluators. Next, we categorize the existing LLM-UM techniques based on their applications. Finally, we outline the remaining challenges and future directions in the LLM-UM field. This survey serves as a handbook for LLM-UM researchers and practitioners to study and use LLMs to augment user modeling systems, and inspires additional interest and work on this topic.

\small{
\bibliographystyle{abbrvnat}
\bibliography{ref}
}


\end{document}